\documentclass[lettersize,journal]{IEEEtran}
\usepackage{amsmath,amsfonts}
\usepackage{algorithmic}
\usepackage{algorithm}
\usepackage{array}
\usepackage{textcomp}
\usepackage{stfloats}
\usepackage{url}
\usepackage{verbatim}
\usepackage{graphicx}
\usepackage{cite}
\usepackage{caption}
\usepackage{calc}

\usepackage{amsmath}  % 等式
\usepackage{pifont}   % 字符
\usepackage{multirow} % 表格
\usepackage[table,xcdraw]{xcolor} % 用于颜色设置
\usepackage{colortbl}             % 用于表格背景色
\usepackage{subcaption}

\usepackage{tikz}
\usetikzlibrary{backgrounds}

\definecolor{IceBlue}{rgb}{0.88,0.95,1.0}
\definecolor{CalmBlue}{rgb}{0.80,0.88,0.98}
\definecolor{AcademicBlue}{rgb}{0.75,0.85,0.95}
\definecolor{SlateBlue}{rgb}{0.70,0.78,0.90}
\definecolor{SkyMist}{rgb}{0.85,0.93,0.98}

\definecolor{MintGreen}{rgb}{0.88,1.0,0.94}
\definecolor{SoftTeal}{rgb}{0.78,0.92,0.88}
\definecolor{PaleGreen}{rgb}{0.86,0.95,0.86}
\definecolor{AquaGreen}{rgb}{0.80,0.90,0.85}
\definecolor{NatureGreen}{rgb}{0.74,0.86,0.78}

\definecolor{CreamYellow}{rgb}{1.0,0.98,0.85}
\definecolor{SoftGold}{rgb}{0.98,0.92,0.70}
\definecolor{LightAmber}{rgb}{1.0,0.94,0.80}
\definecolor{WarmSun}{rgb}{0.96,0.90,0.75}
\definecolor{Apricot}{rgb}{1.0,0.92,0.85}

\definecolor{Lavender}{rgb}{0.93,0.90,0.98}
\definecolor{LightLilac}{rgb}{0.95,0.88,0.98}
\definecolor{SoftPink}{rgb}{1.0,0.88,0.92}
\definecolor{RoseMist}{rgb}{0.98,0.90,0.94}
\definecolor{BlushPink}{rgb}{0.97,0.85,0.88}

\definecolor{NatureBlue}{rgb}{0.36,0.54,0.66}
\definecolor{NatureOrange}{rgb}{0.83,0.56,0.33}
\definecolor{NaturePurple}{rgb}{0.65,0.48,0.62}
\definecolor{NatureGray}{rgb}{0.70,0.70,0.70}

\definecolor{LightGray}{rgb}{0.93,0.93,0.93}
\definecolor{SoftBlue}{rgb}{0.88,0.94,1.0}
\definecolor{SoftGreen}{rgb}{0.88,1.0,0.88}
\definecolor{SoftYellow}{rgb}{1.0,1.0,0.88}
\definecolor{SoftPink}{rgb}{1.0,0.88,0.88}
\definecolor{PaleOrange}{rgb}{1.0, 0.95, 0.85} % 浅橙色
\definecolor{SkyBlue}{rgb}{0.8, 0.9, 1.0}      % 天蓝色

\usetikzlibrary{patterns}
\usepackage{tabularx}

\newcommand{\slashrowcellfixed}[2][\normalsize]{%
  \tikz[baseline=(X.base)]{
    \node[
      minimum height=1.8em,
      inner sep=1pt,
      text width=\linewidth,      % 在 multicolumn 内，等于整行宽
      align=center,
      pattern=north east lines,
      pattern color=red!30,
    ] (X) {};
    
    \node[
      minimum height=1.8em,
      inner sep=1pt,
      text width=\linewidth,
      align=center
    ] at (X.center) {#1 #2};
  }%
}

\newcommand{\gridcell}[2][\normalsize]{% #1=字体命令, #2=文字
  \begin{tikzpicture}[baseline=(X.base)]
    \begin{scope}[on background layer]
      \node[
        minimum height=1.8em,
        inner sep=0pt,
        text width=\linewidth,
        pattern=north east lines,
        pattern color=red!10,
        align=center
      ] (Xbg) {};
      \node[
        minimum height=1.8em,
        inner sep=0pt,
        text width=\linewidth,
        pattern=north west lines,
        pattern color=red!20,
        align=center
      ] at (Xbg.center) {};
    \end{scope}

    \node[
      minimum height=1.8em,
      inner sep=0pt,
      text width=\linewidth,
      align=center
    ] (X) {\centering #1 #2};
  \end{tikzpicture}%
}

\begin{document}

\title{Condition Weaving Meets Expert Modulation: Towards Universal and Controllable Image Generation}

\author{Guoqing Zhang, Xingtong Ge,~\IEEEmembership{Student Member,~IEEE}, Lu Shi,~\IEEEmembership{Student Member,~IEEE}, Xin Zhang, Muqing Xue, Wanru Xu$^*$ ,~\IEEEmembership{Member,~IEEE}, Yigang Cen$^*$,~\IEEEmembership{Member,~IEEE}, Yidong Li,~\IEEEmembership{Senior Member,~IEEE}
\thanks{Guoqing Zhang, Lu Shi, Wanru Xu, Yigang Cen, Yidong Li are with the State Key Laboratory of Advanced Rail Autonomous Operation, the School of Computer Science and Technology, and Visual Intellgence +X International Cooperation Joint Laboratory of MOE, Beijing Jiaotong University(email: guoqing.zhang@bjtu.edu.cn, ygcen@bjtu.edu.cn)} 
\thanks{Xingtong Ge is with the Hong Kong University of Science and Technology}
\thanks{Xingtong Ge and Xin Zhang are with the SenseTime Research}
\thanks{Muqing Xue is with the Beijing Jiaotong University}
\thanks{This work was done by Guoqing Zhang during internship at SenseTime Research Institute.}
\thanks{$*$ Corresponding author.}}

\markboth{Journal of \LaTeX\ Class Files,~Vol.~14, No.~8, August~2021}%
{Shell \MakeLowercase{\textit{et al.}}: A Sample Article Using IEEEtran.cls for IEEE Journals}


\maketitle

\begin{abstract}
The image-to-image generation task aims to produce controllable images by leveraging conditional inputs and prompt instructions. However, existing methods often train separate control branches for each type of condition, leading to redundant model structures and inefficient use of computational resources. To address this, we propose a \textbf{Uni}fied image-to-image \textbf{Gen}eration \textbf{(UniGen)} framework that supports diverse conditional inputs while enhancing generation efficiency and expressiveness.
Specifically, to tackle the widely existing parameter redundancy and computational inefficiency in controllable conditional generation architectures, we propose the \textbf{Condition Modulated Expert (CoMoE)} module. This module aggregates semantically similar patch features and assigns them to dedicated expert modules for visual representation and conditional modeling. By enabling independent modeling of foreground features under different conditions, CoMoE effectively mitigates feature entanglement and redundant computation in multi-condition scenarios. Furthermore, to bridge the information gap between the backbone and control branches, we propose \textbf{WeaveNet}, a dynamic, snake-like connection mechanism that enables effective interaction between global text-level control from the backbone and fine-grained control from conditional branches.
Extensive experiments on the Subjects-200K and MultiGen-20M datasets across various conditional image generation tasks demonstrate that our method consistently achieves state-of-the-art performance, validating its advantages in both versatility and effectiveness. The code has been uploaded to \url{https://github.com/gavin-gqzhang/UniGen}.
\end{abstract}


\section{Introduction}
\label{sec:intro}

\begin{figure}
    \centering
    \includegraphics[width=\linewidth]{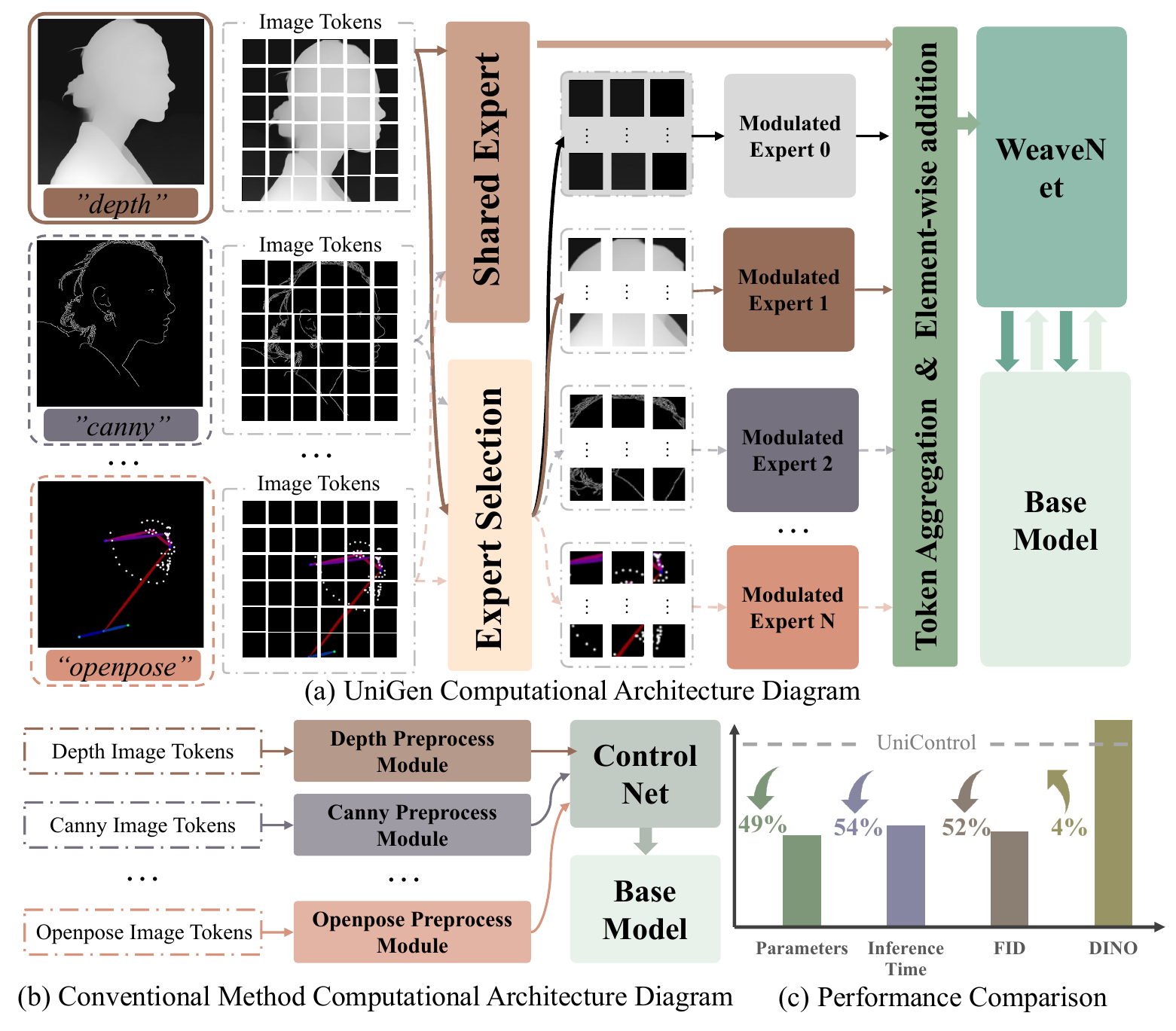}
    \caption{(a) Overall architecture of UniGen, which employs Modulated Experts for conditional modulation and feature alignment to reduce redundant computation, and integrates textual and visual control signals through the interactive WeaveNet design. (b) MoE-based architectures used in related methods \cite{Unicontrol}, where separate preprocessing units are designed for each condition. (c) Performance comparison showing UniGen’s improvements over UniControl \cite{Unicontrol} in parameters, inference speed, and evaluation metrics.}
    \label{fig:motivation}
\end{figure}

With the continuous advancement of diffusion-based methods \cite{ddpm,improved_ddpm,stable_diffusion,sd35,flux.1}, model architectures have evolved from the early UNet \cite{unet,ddpm} design to the more recent DiT \cite{dit} structure, leading to significant improvements in the quality of synthesized images, particularly in text-guided image generation tasks. However, relying solely on text-to-image generation often fails to meet the requirements for controllable synthesis in scenarios demanding precise spatial or structural guidance. To address this limitation, image-to-image generation based on conditional visual constraints has emerged as another important branch of image synthesis. This task aims to generate images under spatial constraints using various conditional inputs, such as depth, canny, and pose information, etc.

To tackle this task, ControlNet \cite{controlnet} introduces a separate branch to inject spatial condition information into the generation process. IP-Adapter \cite{ipadapter} incorporates an adapter structure to enable partial parameter tuning while enforcing spatial constraints. Subsequently, several works \cite{Unicontrol,Pixelponder} have explored approaches inspired by mixture-of-experts (MoE) architectures to handle diverse types of visual conditions. More recently, with the rapid development of the FLUX \cite{flux.1} model in image synthesis, a number of studies \cite{Unicombine,ominicn,Easycontrol} have adopted lightweight fine-tuning techniques such as LoRA \cite{lora} to enhance spatial constraint capabilities from conditional images by tuning only a subset of model parameters.

However, existing methods—whether through constructing MoE architectures or introducing additional parameter tuning—typically design independent control units or control branches for each conditional visual input. Therefore, two major limitations remain in diverse conditional image generation tasks: \textbf{Firstly, redundant parameters, inefficient computation and information forgetting under multi-condition control.} As shown in Fig. \ref{fig:motivation} (b), some methods \cite{Unicontrol} adopt an MoE-like design \cite{moe_1} that selects specific preprocessing modules solely based on the input condition type. This differs substantially from mixture-of-experts architectures, which are designed for efficient parallel computation. In addition, certain approaches \cite{ominicn} build independent and parallel control units for each condition, resulting in severe parameter redundancy. These methods share a fundamental limitation: they overlook the substantial differences between various conditions as well as the underlying commonalities among them. These approaches rely heavily on condition-specific designs, where each condition type is assigned an independent processing module for feature integration. Such designs introduce substantial parameter redundancy and greatly reduce computational efficiency, thereby constraining the model’s practical utility.
Moreover, using isolated modules for each condition type prevents effective alignment of information when handling tasks involving multiple conditions. The model struggles to aggregate heterogeneous conditional cues and suffers from severe information-forgetting problems, making it difficult to establish meaningful correlations among multiple image-level conditions. As a result, the model fails to achieve coherent joint control across conditions, which significantly limits its scalability and generalization ability.
\textbf{Secondly, a lack of effective interaction between the global visual representations guided by text and the local visual representations guided by conditional images.} Existing methods typically rely on isolated conditional control modules to extract conditional visual representations, failing to exploit the dynamic interplay between prompt-guided global (holistic) features and condition-image-guided local (spatial) features. This lack of interaction introduces semantic discrepancies between global and local visual cues, often leading to visually abrupt or inconsistent synthesis results.

To address the above challenges, we propose a unified image generation framework, as illustrated in Fig. \ref{fig:motivation} (a).
Firstly, to address the significant computational redundancy caused by existing methods that employ separate modules or branches for processing different conditional visual representations, we propose the Condition Modulated Expert (CoMoE) module. Our method abandons the traditional paradigm in which independent, non-interfering modules are designed for different condition types. Instead, we start from the common nature of conditional visual representations and focus on the shared foreground-related information across visual tokens. By uncovering and aggregating the latent commonalities among different visual tokens, we employ a conditional modulated expert to fuse token-level visual features with semantic conditions. Compared with traditional approaches, our method not only supports high-quality image synthesis under single-condition control, but also extends easily and effectively to multi-condition joint control, enabling high-quality image generation under multiple conditioning inputs. Moreover, as shown in Fig. \ref{fig:motivation}(c), compared with UniControl \cite{Unicontrol}, which adopts an MoE-like architecture, our method significantly improves image quality while reducing both parameter count and inference latency by approximately 50\%. At the same time, it enables joint control under multiple input conditions. 
Furthermore, to bridge the semantic gap between global visual representations of the base model and local spatial representations from the conditional control units, caused by their independent processing, we propose the WeaveNet architecture. By employing a streamlined serpentine data flow, WeaveNet effectively integrates prompt-guided global semantic constraints with condition image-guided local spatial constraints, enhancing their interaction and significantly mitigating discrepancies arising from separately processing prompts and conditional images.
To validate the effectiveness of our approach, we conduct experiments on the MultiGen-20M \cite{cn++} and Subjects-200K \cite{ominicn} datasets. Compared to both traditional and latest methods, our approach achieves superior performance across multiple evaluation metrics, with only a slight performance gap in certain metrics compared to methods built upon the FLUX \cite{flux.1} base model.

The main contributions of this paper are summarized as follows:
\begin{itemize}
    \item A unified generative framework, dubbed \textbf{UniGen}, is proposed to support controllable image synthesis under arbitrary types of conditional inputs.
    
    % \item The \textbf{Condition Modulated Expert (CoMoE)} module is introduced to enhance foreground feature constraints by leveraging the similarity of local visual representations (patch features) in the conditional image, thereby avoiding redundant parameters and resource overhead caused by designing separate preprocessing modules for each condition.
    \item The \textbf{Condition Modulated Expert (CoMoE)} module is proposed to capture both the differences and commonalities among various types of conditional visual representations. It aggregates highly similar token-level features into dedicated expert modules for efficient information fusion, thereby reducing parameter redundancy and computational overhead.

    \item The \textbf{WeaveNet} architecture is proposed to model the dynamic interaction between the global visual information guided by text and the local spatially visual information from the conditional image, addressing the semantic gap and inconsistent representation caused by independently processing global and local visual cues.
\end{itemize}

\section{Related Works}
\label{sec:relate_work}

\subsection{Diffusion Models}
With the rapid development of diffusion models in the field of image generation, they have gradually emerged as a new generative modeling paradigm following GANs. Initially proposed by \cite{diffusion_model}, diffusion models add noise to original data through a forward process and learn the reverse process to reconstruct the original image. Denoising Diffusion Probabilistic Models (DDPM) \cite{ddpm} significantly improved sampling stability and generation quality. To further accelerate the sampling process and enhance output quality, \cite{ddim} introduced a non-Markovian diffusion process. \cite{improved_ddpm} incorporated multiple training strategies to boost model performance. SDE-based diffusion models \cite{SDE} unified the diffusion framework from the perspective of stochastic differential equations and proposed gradient-based sampling strategies, enhancing model flexibility.
In recent years, with the widespread adoption of Transformer architectures in diffusion models, models such as DiT \cite{dit} have replaced the traditional U-Net \cite{unet} backbone and achieved remarkable progress in high-resolution image generation tasks. Furthermore, the emergence of the Stable Diffusion \cite{stable_diffusion,sd35} and FLUX \cite{flux.1} models has led to substantial improvements in text-controlled image generation. However, considering that the FLUX series involves larger parameter sizes and higher complexity than the Stable Diffusion models, we adopt Stable Diffusion as the backbone of our proposed UniGen architecture to achieve controllable image generation with greater efficiency.

\subsection{Controllable Image Generation}
Controllable image generation aims to guide image synthesis by jointly leveraging conditional images and textual prompts. Existing approaches can be broadly categorized into two groups. The first category focuses on generating semantically aligned content with the text prompt, without requiring strict spatial alignment with the conditional image. For example, \cite{ipadapter} introduces an adapter mechanism that aggregates visual and textual inputs to jointly constrain the generation process. \cite{blip_diff} leverages a multimodal vision-language model to fuse conditional images and semantic information, which are then fed into a diffusion model for controllable generation. \cite{dreambooth,dreamrender} employs a Multi-Modal Attention (MM Attention) mechanism to enable targeted condition injection and personalized image synthesis.
The second category emphasizes preserving both semantic and spatial consistency with the conditional image. Early works such as \cite{controlnet} achieve spatial alignment by encoding conditional information through a parallel branch and integrating it with the text-guided visual features from the backbone network. \cite{Unicontrol} extends this idea by designing a Mixture-of-Experts (MoE)-like architecture \cite{moe_1,moe_2} that supports controllable generation from diverse input conditions. More recently, FLUX \cite{flux.1} has emerged as a strong backbone for text-controlled image generation. Several works \cite{ominicn,Unicombine,Easycontrol} fine-tune FLUX with LoRA \cite{lora} modules on different conditional image types to achieve controllable generation. \cite{Pixelponder} takes a different approach by introducing an MoE Adapter and Patch Adaption module to align multi-condition visual features at the preprocessing stage.

Although these methods demonstrate strong controllability, they often rely on training multiple LoRA \cite{lora} or ControlNet \cite{controlnet} structures or designing parallel preprocessing modules for visual condition alignment. This increases both the number of control parameters and model complexity, significantly limiting generalization. Moreover, independently handling each condition leads to redundant parameters and ignores shared representations among conditional inputs. To address these challenges, this paper analyzes the commonalities and distinctions among visual representations of different condition types and proposes an approach that aggregates similar features using independent modules to reduce redundant computation. Additionally, an interactive injection strategy is introduced to align globally text-guided visual features with locally condition-guided features, thereby bridging the semantic gap between them.

\section{Method}
\label{sec:method}

\begin{figure*}[h]
    \centering
    \begin{subfigure}[b]{0.55\linewidth}
        \centering
        \includegraphics[width=\textwidth]{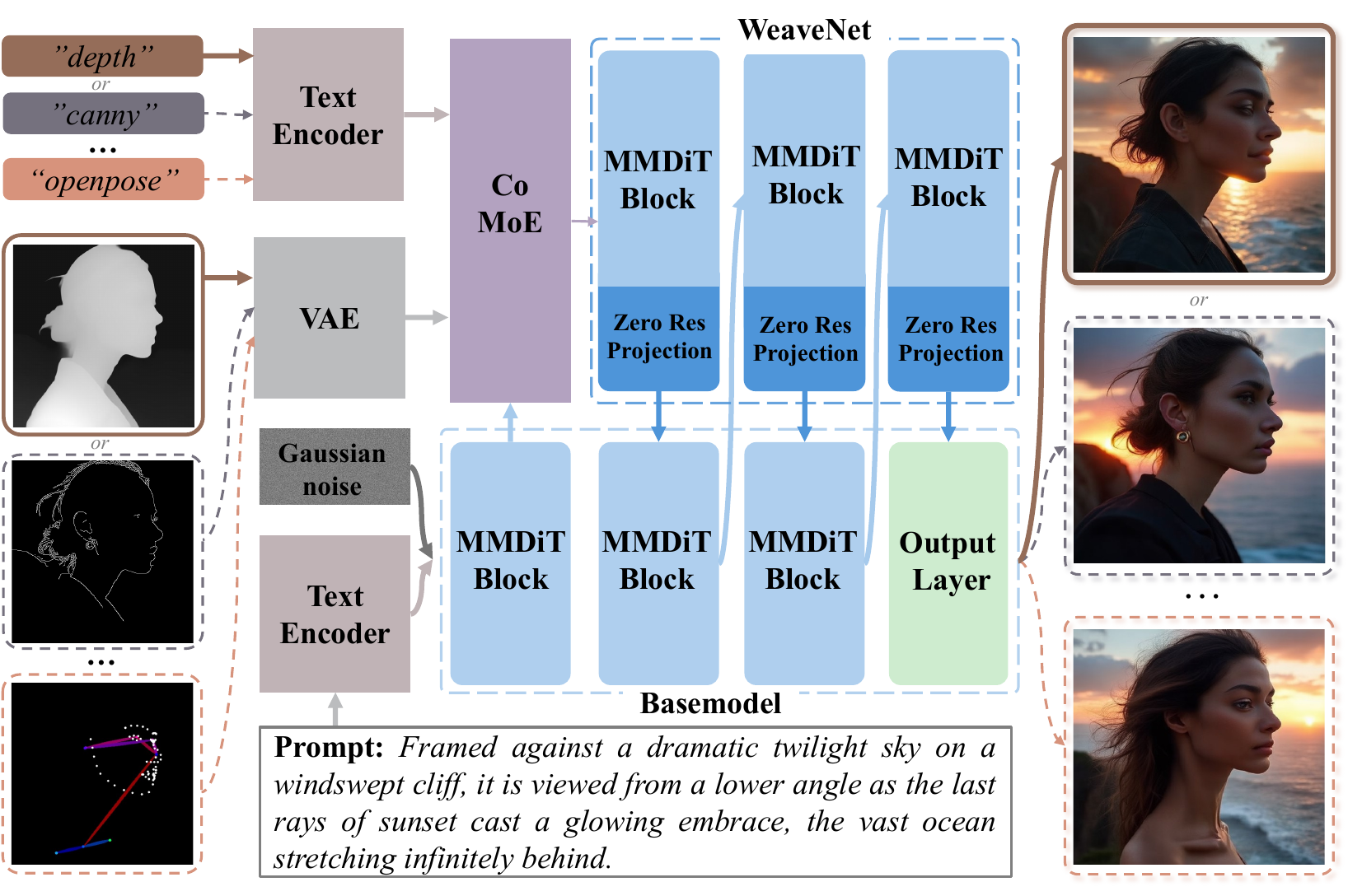}
        \caption{The UniGen architecture supports image inputs with arbitrary types of conditioning to control the content of image generation.}
        \label{fig:model_overview}
    \end{subfigure}
    \hfill
    \begin{subfigure}[b]{0.42\linewidth}
        \centering
        \includegraphics[width=\textwidth]{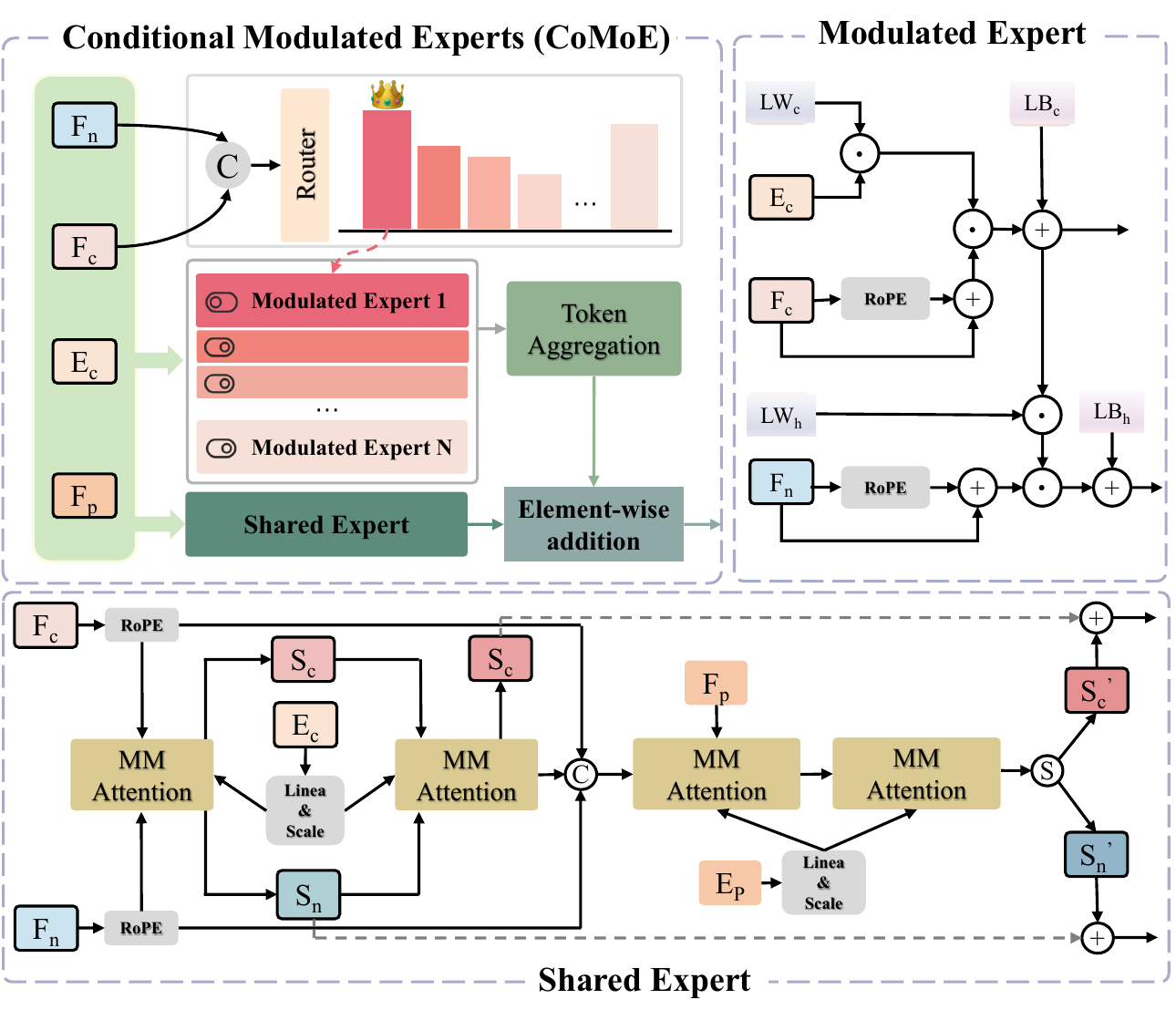}
        \caption{The architecture of Condition Modulated Experts (CoMoE) Module.}
         \label{fig:comoe}
    \end{subfigure}
    \caption{(a) Overall architecture of UniGen, enabling controllable image generation with arbitrary conditional inputs. (b) Structure of Condition Modulated Experts (CoMoE), which modulates and aligns features based on conditional information, achieving parameter-efficient and high-quality controllable generation through a unified branch.}
    % \vspace{-15pt}
    \label{fig:method}
\end{figure*}

\begin{figure}
    \centering
    \includegraphics[width=\linewidth]{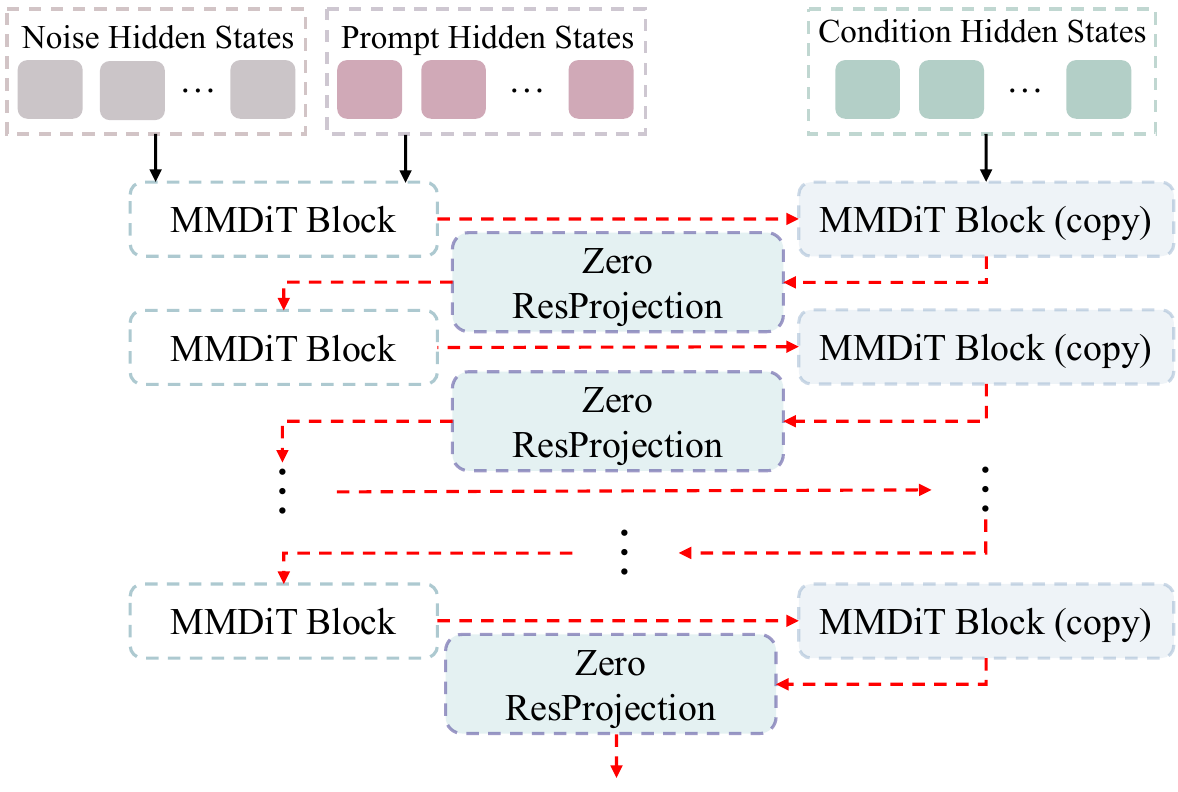}
    \caption{WeaveNet architecture diagram: conditional control information is injected by interleaving the backbone network with conditional-control units.}
    \label{fig:weavenet}
\end{figure}

To better support visual control under diverse conditional inputs and enable a unified image-to-image generation framework, we propose UniGen. As illustrated in Fig. \ref{fig:model_overview}, UniGen accommodates image generation guided by any of 12 types of conditional inputs. During training, it follows the computational framework of existing image-to-image generation methods \cite{ominicn,Unicontrol} and performs conditional noise prediction using a Flow Matching-based \cite{flow_matching} approach.
First, to address the computational and parameter redundancy caused by existing methods that employ parallel and independent modules for different condition types, we introduce the Condition Modulation Expert (CoMoE) module. As shown in Fig. \ref{fig:comoe}, we leverage the commonalities and differences across diverse condition types to guide expert module selection based on the similarity between patch representations. For patches within the same condition type, we independently process the similar representations of foreground regions, aiming to minimize parameter redundancy.
Second, to tackle the lack of effective interaction between global visual representations guided by text and local spatial visual representations guided by conditional images, we introduce the WeaveNet architecture. As illustrated in Fig. \ref{fig:weavenet}, WeaveNet enhances interactive control between text and conditional images through a staggered serpentine computation flow, mitigating local semantic discrepancies and inconsistencies.

\subsection{Condition Modulated Experts (CoMoE)}
\label{sec:comoe}
To fully exploit the diversity and consistency among various types of conditional visual representations while reducing computational and parameter redundancy, we propose a Conditional Modulation Expert (CoMoE) module. As illustrated in Fig. \ref{fig:comoe}, inspired by the design of Mixture of Experts (MoE) commonly used in large multimodal models \cite{moe_llava, remoe}, the module is composed of independent experts and shared experts. The independent experts are designed to capture common foreground information within the same type of conditional representations, enabling joint representation learning of conditional semantics and visual features for condition-specific visual preprocessing. The shared experts focus on integrating global visual representations guided by textual prompts with local conditional visual features, ensuring consistency between condition information and prompt guidance while reducing their semantic gap.

\paragraph{Modulated Expert}
First, to address representation ambiguity, information loss, and misalignment caused by certain sparsely structured visual conditions (e.g., Canny, HED, HEDSketch), we introduce a text-guided global visual feature $F_n$. Note that $F_n$ denotes the features after the first Transformer Block of the backbone network(DiT), where an initial stage of vision–text fusion has already been performed. By fusing the global representation $F_n$ with the condition-specific representation $F_c$, we jointly determine the selection of independent expert modules, allowing better control over the denoising direction, as shown in Equation \ref{eq:pre_expert}. Both $F_n$ and $F_c$ are preprocessed by a VAE \cite{vae} and then converted into token-level visual representations using patch embedding. During training, noise is added to the encoded visual representations $F_n$ with a strength factor $\sigma$ as follows: $F_n=\sigma \cdot N + (1 - \sigma) \cdot F_n$, where $N$ denotes the noise. During inference, $F_n = N$ is used directly as the input to the diffusion model.

\begin{equation}
\begin{cases}
      S_e=Linear(F_n + F_c)), \\
      I_d=Max(S_e, dim=-1)
\end{cases}
\label{eq:pre_expert}
\end{equation}

In the above equation, $S_e$ denotes the expert scores predicted by fusing global visual representations with condition-specific visual representations. $I_d$ represents the indices of expert modules assigned to different patch tokens based on these scores. Guided by these indices, each expert module independently processes foreground regions with higher feature similarity, while background regions are grouped and handled by a separate expert module.

Secondly, when performing subsequent independent processing of vision-level tokens, all visual tokens are first dispersed and then passed to separate Modulated Expert modules based on their feature correlations. This procedure, however, may cause the loss of global structural and spatial information from the original conditioning image. Therefore, before discretizing the tokens, we introduce Rotary Positional Embeddings (RoPE) \cite{rope} to encode spatial positional information, ensuring spatial consistency and preventing spatial confusion in the aggregated visual features after independent expert processing (we provide experimental analysis of RoPE in Section \ref{sec:ablation_moe_rope}). In addition, within the Modulated Expert module, our goal is to leverage condition-specific semantics to process local visual representations and fuse them with spatially aligned, text-guided global visual features.
Specifically, we construct a learnable parameter matrix $LW_c$, and extract the semantic representation $E_c$ based on the category of the given condition. The condition-aware semantic embedding is then used to modulate the parameter matrix, enabling one distinct matrix per condition for conditional feature representation, as shown in Equation \ref{eq:condition_modulated}.

\begin{equation}
\begin{cases}
    F_n^{'} = RoPE(F_n), F_{c}^{'} = RoPE(F_c),\\
    E_c = Pooling(CLIP([condition\_types])), \\
    LW_c \sim \mathcal{N}(0, 0.1^2), F_{c}^{'} = LW_c \cdot E_c \cdot F_{c}^{'}, \\
    LB_c \sim \mathcal{N}(0, 0.1^2), F_{c}^{'} = F_{c}^{'} + LB_c
\end{cases}
\label{eq:condition_modulated}
\end{equation}

Meanwhile, to further enhance the ability of condition-specific visual representations to constrain the global visual features guided by prompts, we follow the same design principle. A learnable parameter matrix $LW_h$ is constructed and modulated using the preprocessed conditional visual feature $F_c^{'}$ in Equation \ref{eq:condition_modulated} to reconstruct the feature mapping parameters. This modulation allows control over the effective regions of the global visual representation. During this process, the use of rotary position encoding ensures strict spatial alignment between the representations, thereby avoiding spatial information confusion.

\begin{equation}
\begin{cases}
    LW_h \sim \mathcal{N}(0, 0.1^2), F_{n}^{'} = LW_h \cdot F_{c}^{'} \cdot F_n^{'}, \\
    LB_h \sim \mathcal{N}(0, 0.1^2), F_{n}^{'} = F_{n}^{'} + LB_h
\end{cases}
\label{eq:noised_modulated}
\end{equation}

Finally, after aligning the visual representation of the condition with the global visual representation guided by the prompt, each output from the modulated experts corresponds to the same spatial location and semantic type. This alignment effectively enhances the constraint imposed by the condition on the global visual representation. Therefore, we perform reverse aggregation based on the visual token indices $I_d$ computed in Equation \ref{eq:pre_expert} to reconstruct the original ordered distribution of visual tokens.

\begin{equation}
\begin{cases}
    \hat{F_n} \sim {Zero}(F_n), \hat{F_c} \sim {Zero}(F_c), \\
    \hat{F_n} = Reverse(F_n^{'}, \hat{F_n}, I_d), \hat{F_c} = Reverse(F_c^{'}, \hat{F_c}, I_d)
\end{cases}
\label{eq:noised_modulated}
\end{equation}

In the above equation, $Reverse$ refers to reordering the same-type visual representations computed by the Modulated Expert based on the index $I_d$ obtained from Equation \ref{eq:pre_expert}, thereby restoring the original distribution of image tokens.

Although this design aims to aggregate homogeneous conditional visual information and perform directional, condition-guided feature enhancement to achieve discretized feature alignment, it inevitably suffers from a common issue in expert-based architectures: a few Modulated Expert modules may be overly activated, dominating the modulation of most visual tokens. This results in dense computation despite the intended sparsity. Meanwhile, our Modulated Expert modules are designed to fuse and process visually similar tokens based on conditional semantic representations. Excessive activation of a single Modulated Expert increases training difficulty and harms convergence. To address this, we introduce an expert regularization loss \cite{moe_loss} widely used in large language models. By minimizing the inner product between expert scores and expert activation counts, the loss prevents the router from over-assigning tokens to a small subset of experts, thereby encouraging balanced expert utilization.
In addition, this regularization effectively prevents visual tokens from being overly routed to a single Modulated Expert, ensuring that each expert focuses on a more coherent subset of visual representations and accelerating convergence during training.

\paragraph{Shared Expert}
The shared expert module leverages a traditional Multi-Modal Attention(MMAttention) mechanism to achieve conditional image generation from two perspectives: one guided by the condition's visual representation for directional denoising, and the other by the prompt for global denoising. First, the visual representation of the condition is enhanced by encoding its semantic embedding $E_c$ together with the selected diffusion timestep, which serves as a basis for feature filtering and scaling. Then, attention is used to compute the feature correlation between the condition’s visual features $F_{c}$ and the prompt-guided global representation $F_{n}$, thereby reinforcing the influence of the condition in the global visual context. Finally, an attention mechanism is employed to enhance the correlation between local regions within the visual representation of the condition.

\begin{equation}
\begin{cases}
    T_n = Embedding(timestep,E_c),  \\
    S_n, S_c = MMAttention(RoPE(F_n), RoPE(F_c), T_n), \\
    S_c = MMAttention(S_c, S_c, T_n)
\end{cases}
\label{eq:se1_modulated}
\end{equation}

% In the above equation, RoPE denotes Rotary Positional Encoding, which aims to achieve spatial alignment between conditional visual representations and prompt-guided global visual representations, thereby preventing the failure of localized constraints in the conditional visual features. The embedding layer jointly encodes the pooled features of the condition and the timestep, and the resulting embeddings serve as input to the MMAttention module for computing the offset and scale of the attention-refined features.
In the above equation, embedding layer jointly encodes the pooled features of the condition and the timestep, and the resulting embeddings serve as input to the MMAttention module for computing the offset and scale of the attention-refined features.

% Moreover, to better integrate the joint constraints of the condition and prompt, we perform a joint embedding of the prompt's pooled features $E_p$ and the timestep to obtain a new representation $T_n^{'}$, which is used to modulate the shift and scale in the MMAttention module. We employ prompts to optimize both global and condition-specific visual representations, ensuring accurate instruction adherence under each representation. 
Furthermore, to better integrate the joint constraints of conditional visual representations and text prompts, we adopt a similar design strategy as described above to implement the shift and scale operations in the MMAttention module. Meanwhile, we employ text prompts to refine both global and condition-specific visual representations, ensuring accurate instruction adherence under each representation. Through the use of an attention mechanism, the prompt not only guides the global visual representation but also influences the condition representation, thereby avoiding the independence constraint between the prompt and condition representations, as shown in Equation \ref{eq:se2_modulated}.

\begin{equation}
\begin{cases}
    T_n^{'} = Embedding(timestep,E_p),  \\
    S_n^{'}, S_c^{'} = MMAttention([RoPE(F_n), RoPE(F_c)], F_p, T_n^{'}), \\
    S_n^{'}, S_c^{'} = MMAttention([S_n^{'}, S_c^{'}], [S_n^{'}, S_c^{'}], T_n^{'}), \\
    S_n^{'} = S_n + S_n^{'}, S_c^{'} = S_c + S_c^{'}
\end{cases}
\label{eq:se2_modulated}
\end{equation}

In the above equation, $[\cdot, \cdot]$ denotes the token-level concatenation of features. $F_p$ denotes the semantic representation of the prompt encoded by CLIP \cite{clip} and T5 \cite{t5}.

Finally, we aggregate the condition-aware visual representations computed from both the Modulated Expert and the Shared Expert, and also combine the globally constrained visual representations of the condition to enhance spatial-level feature consistency.

\begin{equation}
\begin{cases}
    \hat{F_n} = \hat{F_n} + S_n^{'}, \\
    \hat{F_c} = \hat{F_c} + S_c^{'},
\end{cases}
\label{eq:add_modulated}
\end{equation}

\subsection{WeaveNet}
\label{sec:weavenet}

To address the semantic gap between textual prompts and conditional visual control introduced by existing methods such as ControlNet \cite{controlnet}, which processes conditional visual features and textual guidance in parallel via independent branches, we propose a novel architecture named WeaveNet. As illustrated in Fig. \ref{fig:weavenet}, WeaveNet introduces an interactive conditional information injection mechanism that enables dynamic interaction between conditional visual representations and text-guided visual features.
Specifically, following the architectural design of ControlNet, we replicate the modules and parameters of the base model to construct a separate branch for processing conditional inputs. Unlike ControlNet, WeaveNet integrates features from the base model into the corresponding layers of the control branch, effectively combining the global semantics from textual prompts with the local information from conditional images. Then, the Zero ResProjection Module is employed to align the outputs of each layer in the control branch and inject them into the main network, ensuring dynamic interaction between the prompt and the conditional image. This helps prevent semantic gaps caused by overemphasis on either side. The detailed computational process is presented in Algorithm \ref{alg:weavenet}.

\begin{algorithm}[h!]
\caption{The computational workflow of the WeaveNet architecture.}
\label{alg:weavenet}
\begin{algorithmic}[1] 
\STATE init $WeaveBlocks = copy(DiTBlocks)$  
\STATE init $F_n$, $F_c$  \COMMENT{Noisy visual features and Conditioned visual features obtained via VAE-based encoding}
\STATE init $F_p$, $E_c$, $E_p$  \COMMENT{$F_p$ denotes the semantic embedding representation of the prompt, while $E_c$ and $E_p$ represent the pooled semantic embeddings of the condition type and the prompt, respectively.}
\FOR{index in $range(len(DiTBlocks))$}
    \STATE $F_n = DiTBlocks[index](F_{n},F_p, E_p)$ 
    \IF{index is 0}
        \STATE $\hat{F_n}, \hat{F_c} = CoMoE(F_n,F_c, E_c, E_p, F_p)$ \\ 
        $\hat{F_n} = \hat{F_n} + \hat{F_c} $ \\
        $\hat{F_n} = WeaveBlocks[index](\hat{F_n},F_p, E_c)$
    \ELSE
        \STATE $\hat{F_n} = WeaveBlocks[index](F_n,F_p, E_c)$
    \ENDIF

    \STATE $F_{n}=F_{n}+ ZeroProjection(\hat{F_n})$
\ENDFOR
\STATE \textbf{return} $F_n^{'} = Reshape(Projection(F_n))$ 
\end{algorithmic}
\end{algorithm}

\section{Experiments}
\label{sec:experiments}
\subsection{Datasets}
We conduct training and evaluation on the MultiGen-20M \cite{cn++} and Subjects-200k \cite{ominicn}datasets. MultiGen-20M contains over 2 million image-description pairs with corresponding conditional images. These conditional images cover 12 types of conditions, such as depth, Canny edges, OpenPose, HED, normals, and grayscale. However, the image descriptions in this dataset exhibit significant inconsistencies and low accuracy, which introduces training bias and degrades generation quality. To address this issue, we re-annotated all captions in the dataset using Qwen\footnote{https://huggingface.co/Qwen/Qwen2.5-VL-7B-Instruct} \cite{qwen}. The newly annotated captions have been uploaded to Hugging Face\footnote{https://huggingface.co/datasets/gavin-zhang/MultiGen20M\_json}. Furthermore, to ensure fair and accurate comparisons, all methods are trained and evaluated using the same training and test splits. Compared with MultiGen-20M, the Subjects-200k dataset contains over 200k high-quality images with detailed and accurate entity-level annotations and reliable textual descriptions, along with GPT-based scoring of the captions. However, Subjects-200k provides relatively weaker support for conditional image configurations. To overcome this limitation, we further extend Subjects-200k by generating three types of conditional inputs: Depth, Canny, and OpenPose. Specifically, we use the pre-trained Depth-Anything\footnote{https://github.com/LiheYoung/Depth-Anything} \cite{depthanything} model to produce depth maps for each image, and the pre-trained DWpose\footnote{https://github.com/IDEA-Research/DWPose} \cite{dwpose} model to extract human keypoints (OpenPose). For the Canny \cite{canny} condition, we generate edge maps online using the OpenCV-Python library. Note that DWpose may incorrectly produce pose information for images without humans. To avoid this, we apply an additional filtering step using Qwen: for each image, we determine whether it contains human subjects, and remove the corresponding OpenPose condition if no humans are present. All extended data have been uploaded to Hugging Face\footnote{https://huggingface.co/datasets/gavin-zhang/Subjects200K}. We also strictly separate the training and test sets and retrain all compared methods to ensure fair and valid evaluation. For both MultiGen-20M and Subjects-200k, the test set consists of 5,000 condition–description–reference image triplets, kept entirely independent from the training data. In addition, since Subjects-200k supports both single-condition control and multi-condition joint control generation tasks, its test set includes samples covering both settings. This design ensures independence from the training set and prevents data leakage, thereby avoiding biased evaluation in conditional control tasks.

\subsection{Evaluation Metrics}
We adopt several widely used evaluation metrics in related work: FID \cite{FID}, SSIM \cite{SSIM}, CLIP-I \cite{clipscore,clip}, CLIP-T \cite{clipscore,clip}, and DINO \cite{DINO}. FID evaluates the realism and distributional similarity between generated and real images at the feature level by extracting features using Inception V3 \cite{inception}. SSIM measures the similarity between two images in terms of luminance, contrast, and structural information. CLIP-I and CLIP-T leverage CLIP to extract visual and semantic representations. The quality of the generated images is assessed by computing feature similarity between generated and real images (CLIP-I), as well as between generated images and text prompts (CLIP-T). DINO also extracts visual features and measures the semantic structural similarity between generated and real images.

To comprehensively evaluate the effectiveness of our method, we additionally incorporate the metrics from ICE-Bench\cite{ice_bench}, covering multiple dimensions including aesthetics, imaging quality, instruction-following ability, reference consistency, and controllability.
From the aesthetics perspective, we adopt AES (Aesthetic Score), which uses an aesthetic predictor to assess the overall visual appeal of generated images. This metric considers composition, color usage, consistency, and naturalness, providing an effective measure of generative quality and visual attractiveness.
For imaging quality, we include IMG (Imaging Quality), which evaluates whether an image is sharp and free of distortions or exposure issues.
To assess instruction adherence, we use CLIP-Cap, which computes the similarity between images and semantic descriptions in the feature space, offering a direct measure of the model’s ability to interpret and follow textual prompts.
Finally, we assess pixel-level differences using L1-based metrics. L1 Raw measures the mean absolute pixel difference between the generated and ground-truth images, reflecting pixel-level alignment. L1 Color evaluates color restoration and color-shift magnitude, indicating the color discrepancy relative to the ground truth. L1 Cond computes the pixel-level difference between the condition image extracted from the generated output and the reference condition image. It reflects how well the generated image spatially fits the reference condition. Following the evaluation protocol of ICE-Bench, all reported values for L1 Color and L1 Cond are presented as $1 - L1_{Color}$ and $1 - L1_{Cond}$, respectively.

To further align with standard evaluation practices in image generation tasks, we extend our assessment using additional image quality metrics. Following the methodology adopted in IQA-Pytorch\footnote{https://github.com/chaofengc/IQA-PyTorch}, we evaluate performance across MUSIQ \cite{musiq}, MAN-IQA \cite{maniqa}, CLIP-IQA \cite{clip-iqa}, LPIPS \cite{lpips}, LAION-Aesthetics \cite{laion_aes}, PSNR, NIMA \cite{nima}, TOPIQ-NR \cite{topiq}, and CLIPScore \cite{clip}. Among these, MUSIQ, CLIP-IQA, LAION-Aesthetics, and CLIPScore share similar purposes with several ICE-Bench metrics, characterizing aesthetic quality, imaging fidelity, and semantic or visual alignment. In addition, we introduce LPIPS, PSNR, NIMA, and TOPIQ-NR, which evaluate perceptual distance, pixel-level error, human aesthetic preference, and no-reference image quality, respectively, enabling a more comprehensive assessment of the generated images.

\subsection{Experimental Setting}
To comprehensively evaluate the effectiveness of our method, we conduct experiments from two perspectives: single-condition control and multi-condition joint control. Since the MultiGen-20M dataset contains 12 types of independent conditioning signals, we use this dataset only for single-condition image generation and compare our method with conventional approaches. The Subjects-200k dataset provides four types of conditioning signals, among which three can be used independently. Therefore, we use the depth, canny, and openpose conditions from Subjects-200k to train and evaluate the single-condition control task. To further assess our method, we follow the experimental design of UniCombine \cite{Unicombine} and construct multi-condition control tasks by combining: Depth + Canny, Subject + Depth, and Subject + Canny. These combinations are used to train and evaluate the joint-condition control setting. In Section \ref{sec:compare_sota_single_condition}, we evaluate and analyze the generation results for single-condition control tasks on the MultiGen-20M and Subjects-200k datasets. In Section \ref{sec:compare_sota_multi_condition}, we focus on evaluating and analyzing the generation results for multi-condition control tasks on the Subjects-200k dataset.

\begin{table*}[t]
% \resizebox{\linewidth}{!}{
\centering
\begin{tabular}{cccccccccc}
\hline
&   & \multicolumn{8}{c}{Condition Type}  \\ \cline{3-10} 
&   & \multicolumn{4}{c}{Subjects-200K} &\multicolumn{4}{c}{MultiGen-20M}  \\
\multirow{-3}{*}{Metrics} & \multirow{-3}{*}{Method}   & Depth & Canny & Openpose & Mean  & Depth  & Canny  & Openpose  & Mean \\ \hline
  & UniControl \cite{Unicontrol}   & 0.33      & 0.42     & 0.24  & 0.33   & 0.19    & 0.19      & 0.19  & 0.19   \\
  & Controlnet ++ \cite{cn++}      & 0.39      & 0.41     & -     & 0.4   & 0.28    & 0.28      & -   & 0.28    \\
  & OminiControl $^\dagger$ \cite{ominicn}   & {0.49}      & \underline{0.58}     & {0.33}   & {0.47}  & \textbf{0.34}    & \textbf{0.43}     & \underline{0.25}  & \underline{0.34}   \\
  & PixelPonder  $^\dagger$ \cite{Pixelponder}    & 0.35   & 0.39  & {0.33}  & 0.36   & 0.28    & 0.33    & \underline{0.25}   & 0.27  \\
& \cellcolor{NatureGreen!30}\textbf{UniGen(ours)} & \cellcolor{NatureGreen!30}\textbf{0.5}  & \cellcolor{NatureGreen!30}{0.55}  & \cellcolor{NatureGreen!30}\textbf{0.39}  & \cellcolor{NatureGreen!30}\underline{0.48}  & \cellcolor{NatureGreen!30}\textbf{0.34}  & \cellcolor{NatureGreen!30}\textbf{0.43}  & \cellcolor{NatureGreen!30}\textbf{0.28}  &  \cellcolor{NatureGreen!30}\textbf{0.35}  \\
\multirow{-6}{*}{SSIM $\uparrow$}    & \cellcolor{NatureGreen!30}\textbf{UniGen(ours)} $^\dagger$ & \cellcolor{NatureGreen!30}\textbf{0.5}  & \cellcolor{NatureGreen!30}\textbf{0.59}  & \cellcolor{NatureGreen!30}\underline{0.37}  & \cellcolor{NatureGreen!30}\textbf{0.49}  & \cellcolor{NatureGreen!30}-  & \cellcolor{NatureGreen!30}- & \cellcolor{NatureGreen!30}-  &  \cellcolor{NatureGreen!30}- \\ \hline
& UniControl  \cite{Unicontrol}      & 23.34      & 18.23  & 36.69  & 26.09  & 57.63 & 57.31 & 57.84   & 57.59   \\
& Controlnet++    \cite{cn++}      & 15.99       & 23.26   & -  & 19.63   & \underline{22.24} & 37.56 & -  & \underline{29.9} \\
& OminiControl $^\dagger$  \cite{ominicn}    & {12.35}     & \underline{9.38}   & 35.03   & {18.92}   & 22.92  & \underline{15.34}  & \underline{52.0}  & 30.09 \\
& PixelPonder $^\dagger$ \cite{Pixelponder}    & 24.11     & 21.86   & {30.98}  & 25.65    & 24.83    & 20.92    & 65.94  & 37.23 \\
& \cellcolor{NatureGreen!30}\textbf{UniGen(ours)} & \cellcolor{NatureGreen!30}\textbf{11.26} & \cellcolor{NatureGreen!30}{9.52}  & \cellcolor{NatureGreen!30}\textbf{15.68} & \cellcolor{NatureGreen!30}\textbf{12.15} & \cellcolor{NatureGreen!30}\textbf{13.45} & \cellcolor{NatureGreen!30}\textbf{11.83} & \cellcolor{NatureGreen!30}\textbf{15.43}    &   \cellcolor{NatureGreen!30}\textbf{13.57}    \\
\multirow{-6}{*}{FID $\downarrow$}     & \cellcolor{NatureGreen!30}\textbf{UniGen(ours)} $^\dagger$ & \cellcolor{NatureGreen!30}\underline{11.51} & \cellcolor{NatureGreen!30}\textbf{8.32}  & \cellcolor{NatureGreen!30}\underline{26.51} & \cellcolor{NatureGreen!30}\underline{15.45} & \cellcolor{NatureGreen!30}-  & \cellcolor{NatureGreen!30}- & \cellcolor{NatureGreen!30}-    &   \cellcolor{NatureGreen!30}-    \\ \hline
 & UniControl    \cite{Unicontrol}  & 84.95  & 87.99   & 77.80 & 83.58   & 64.49   & 64.53  & 64.47  & 64.50 \\
& Controlnet++    \cite{cn++}      & 87.04  & 86.75   & -  & 86.90  & 73.86   & 72.83  & -  & 73.35  \\
& OminiControl$^\dagger$ \cite{ominicn}      & \underline{90.59}  & \underline{93.13}   & {79.84}  & \underline{87.85}   & \underline{75.05} & \underline{82.73}   & 64.9  & \underline{74.23} \\
& PixelPonder$^\dagger$ \cite{Pixelponder}   & 79.98   & 83.35   & 76.89  & 80.07  & 74.17   & 79.50   & \underline{67.64}  & 73.77  \\
& \cellcolor{NatureGreen!30}\textbf{UniGen(ours)} & \cellcolor{NatureGreen!30}{89.14}  & \cellcolor{NatureGreen!30}{91.61} & \cellcolor{NatureGreen!30}\textbf{82.72}  & \cellcolor{NatureGreen!30}{87.82} & \cellcolor{NatureGreen!30}\textbf{76.77} & \cellcolor{NatureGreen!30}\textbf{84.06} & \cellcolor{NatureGreen!30}\textbf{71.54} &  \cellcolor{NatureGreen!30}\textbf{77.46} \\
\multirow{-6}{*}{CLIP-I $\uparrow$}  & \cellcolor{NatureGreen!30}\textbf{UniGen(ours)}$^\dagger$ & \cellcolor{NatureGreen!30}\textbf{90.94}  & \cellcolor{NatureGreen!30}\textbf{93.67} & \cellcolor{NatureGreen!30}\underline{82.47}  & \cellcolor{NatureGreen!30}\textbf{89.03} & \cellcolor{NatureGreen!30}- & \cellcolor{NatureGreen!30}- & \cellcolor{NatureGreen!30}- &  \cellcolor{NatureGreen!30}- \\ \hline
 & UniControl  \cite{Unicontrol}   & 18.49    & 18.24   & 19.18  & 18.64    & 19.23   & 19.54   & 19.38  & 19.38 \\
 & Controlnet++   \cite{cn++}    & 18.83   & 19.36   & -   & 19.10   & 19.33    & 19.95    & -  & 19.64 \\
& OminiControl$^\dagger$  \cite{ominicn}    & {19.96}   & {20.58}   & {20.59}   & {20.38}    & 19.16    & 19.53    & 19.91   & 19.53 \\
& PixelPonder$^\dagger$ \cite{Pixelponder}  & \underline{21.56}  & \underline{21.75}  & \underline{21.4}  & \underline{21.57}   & \textbf{20.62}   & \textbf{20.63}   & \textbf{20.19}   & \textbf{20.48} \\
& \cellcolor{NatureGreen!30}\textbf{UniGen(ours)} & \cellcolor{NatureGreen!30}19.51  & \cellcolor{NatureGreen!30}20.21 & \cellcolor{NatureGreen!30}20.55 & \cellcolor{NatureGreen!30}20.09   & \cellcolor{NatureGreen!30}\underline{19.56}  & \cellcolor{NatureGreen!30}\underline{20.06}  & \cellcolor{NatureGreen!30}\textbf{20.19}  & \cellcolor{NatureGreen!30}\underline{19.94} \\ 
\multirow{-6}{*}{CLIP-T $\uparrow$}  & \cellcolor{NatureGreen!30}\textbf{UniGen(ours)}$^\dagger$ & \cellcolor{NatureGreen!30}\textbf{22.28}  & \cellcolor{NatureGreen!30}\textbf{22.38} & \cellcolor{NatureGreen!30}\textbf{22.21} & \cellcolor{NatureGreen!30}\textbf{22.29}   & \cellcolor{NatureGreen!30}-  & \cellcolor{NatureGreen!30}-  & \cellcolor{NatureGreen!30}-  & \cellcolor{NatureGreen!30}- \\ \hline
& UniControl \cite{Unicontrol}   & 91.23   & 92.35   & {82.28}  & 88.62  & 75.72   & 75.75   & {75.69}  & 75.72  \\
& Controlnet++   \cite{cn++}    & 93.02   & 91.0    & -   & 92.0   & 87.99   & 85.96    & -  & \underline{86.98} \\
& OminiControl $^\dagger$  \cite{ominicn}   & \textbf{95.42}   & \underline{96.14}   & {85.94}  & {92.5}  & \underline{88.39}  & \underline{92.1}   & \underline{79.14}  & 86.54 \\
& PixelPonder$^\dagger$  \cite{Pixelponder}  & 86.03   & 88.97  & 82.79  & 85.93   & 84.78  & 88.48  & 76.22  & 83.16 \\
& \cellcolor{NatureGreen!30}\textbf{UniGen(ours)} & \cellcolor{NatureGreen!30}{94.31}   & \cellcolor{NatureGreen!30}{95.20} & \cellcolor{NatureGreen!30}\underline{88.11} & \cellcolor{NatureGreen!30}\underline{92.54}  & \cellcolor{NatureGreen!30}\textbf{88.68} & \cellcolor{NatureGreen!30}\textbf{92.19} & \cellcolor{NatureGreen!30}\textbf{83.74}  & \cellcolor{NatureGreen!30}\textbf{88.20} \\
\multirow{-6}{*}{DINO $\uparrow$}    & \cellcolor{NatureGreen!30}\textbf{UniGen(ours)}$^\dagger$  & \cellcolor{NatureGreen!30}\underline{95.37}   & \cellcolor{NatureGreen!30}\textbf{96.44} & \cellcolor{NatureGreen!30}\textbf{88.14} & \cellcolor{NatureGreen!30}\textbf{93.32}  & \cellcolor{NatureGreen!30}- & \cellcolor{NatureGreen!30}- & \cellcolor{NatureGreen!30}- & \cellcolor{NatureGreen!30}- \\ \hline
\end{tabular}
\caption{We conduct a comprehensive performance comparison under Depth, Canny, and OpenPose conditions using the Subjects-200K and MultiGen-20M datasets. $\dagger$ denotes the use of FLUX \cite{flux.1} as the base model.}
\label{tab:ss200k}
\end{table*}

\subsection{Implement Details}
We adopt Stable Diffusion \cite{sd35}\footnote{https://huggingface.co/stabilityai/stable-diffusion-3.5-medium} and Flux .1\footnote{https://huggingface.co/black-forest-labs/FLUX.1-schnell} \cite{flux.1} as the base model and initialize it with publicly available pretrained weights. The training framework is implemented using PyTorch Lightning and conducted on 8 A100 GPUs with the standard Distributed Data Parallel (DDP) strategy. When training models based on Flux .1, the parameter count is significantly larger than that of Stable Diffusion. Therefore, we adopt the DeepSpeed ZeRO-3 \cite{deepspeed} configuration provided by Accelerate \footnote{https://github.com/huggingface/accelerate} to partition and offload parameters, which helps reduce the risk of GPU memory overflow.

During training, we use the AdamW \cite{adamw} optimizer with a learning rate of 0.0001, a constant learning rate scheduler with 500 warm-up steps, and a batch size of 16. This work aims to enhance the model's adaptability to various conditioning types. To this end, we optimize the distributed sampler to prevent batches from containing samples of the same conditioning type, which could bias the model’s optimization direction. Similar to other reproduced methods, we randomly sample 160,000 condition-description pairs for each conditioning type from the training set.

For evaluation, we follow the same testing protocol as other reproduced models by randomly sampling 5,000 condition-description pairs per conditioning type from the test set. Inference is performed on a single A100 GPU with a condition scale of 1.0, guidance scale of 3.5, and 28 inference steps. Considering the substantial impact of random seeds on sampling outcomes, we fix the seed to 1872 for all experiments. For potential training-free performance improvements, we suggest exploring the impact of different seeds to assess their adaptability across various prompts.

\begin{table*}[h!]
\centering
\resizebox{\linewidth}{!}{
\begin{tabular}{cccccccccccc}
\hline
       &          & \multicolumn{9}{c}{Condition Types}        \\ \cline{3-11} 
\multirow{-2}{*}{Metric} & \multirow{-2}{*}{Method}             & Hed            & Hedsketch      & Normal         & Seg            & Bbox           & Outpainting    & Inpainting     & Blur           & Grayscale     & \multirow{-2}{*}{Mean} \\ \hline
& UniControl     & 0.19   & 0.19  & 0.19  & 0.19  & 0.19  & 0.19  & 0.19   & 0.19  & 0.19 & 0.19   \\
& Controlnet++  & 0.20   & -   & -   & -  & -   & -   & -  & -   & -     & -      \\
& OminiControl$^\dagger$    & \underline{0.41}  & \underline{0.36}  & \underline{0.36}   & \underline{0.30}  & 0.22   & \textbf{0.56}  & \textbf{0.77}  & \underline{0.70}   & \textbf{0.90} & \underline{0.51}  \\
& PixelPonder$^\dagger$   & 0.33  & 0.32  & 0.29  & 0.27 & \underline{0.25}  & 0.41  & 0.57  & 0.57  & 0.68 & 0.41 \\
\multirow{-5}{*}{\rotatebox{90}{SSIM} \rotatebox{90}{$\xrightarrow{\hspace{0.5cm}}$}}    & \cellcolor{NatureGreen!30}\textbf{UniGen(ours)} & \cellcolor{NatureGreen!30}\textbf{0.45}  & \cellcolor{NatureGreen!30}\textbf{0.41}  & \cellcolor{NatureGreen!30}\textbf{0.37}  & \cellcolor{NatureGreen!30}\textbf{0.31}  & \cellcolor{NatureGreen!30}\textbf{0.26}  & \cellcolor{NatureGreen!30}\textbf{0.56}  & \cellcolor{NatureGreen!30}\underline{0.76}   & \cellcolor{NatureGreen!30}\textbf{0.73}  & \cellcolor{NatureGreen!30}\textbf{0.90}  & \cellcolor{NatureGreen!30}\textbf{0.53} \\ \hline
& UniControl  & 57.51  & 57.55  & 57.81  & 57.78  & 57.72  & 57.79  & 57.78   & 57.87  & 57.88  & 57.74 \\
& Controlnet++  & 46.07  & -   & -   & -   & -   & -  & -   & -  & -   & -  \\
& OminiControl$^\dagger$  & \underline{12.59}   & \underline{20.08}   & \underline{19.56}   & 33.51   & 67.10  & \underline{11.50}   & \textbf{7.46}  & \underline{8.21}    & \underline{7.09}   & \underline{20.79} \\
& PixelPonder$^\dagger$   & 17.26  & 21.87   & 24.15   & \underline{31.24}  & \underline{64.09}   & 19.09  & 14.96   & 13.34   & 11.11   & 24.12 \\
\multirow{-5}{*}{\rotatebox{90}{FID} \rotatebox{90}{$\xleftarrow{\hspace{0.3cm}}$}}     & \cellcolor{NatureGreen!30}\textbf{UniGen(ours)} & \cellcolor{NatureGreen!30}\textbf{8.22}  & \cellcolor{NatureGreen!30}\textbf{12.42} & \cellcolor{NatureGreen!30}\textbf{12.64} & \cellcolor{NatureGreen!30}\textbf{14.19} & \cellcolor{NatureGreen!30}\textbf{16.26} & \cellcolor{NatureGreen!30}\textbf{9.69} & \cellcolor{NatureGreen!30}\underline{7.72}  & \cellcolor{NatureGreen!30}\textbf{7.21}  & \cellcolor{NatureGreen!30}\textbf{6.77} & \cellcolor{NatureGreen!30}\textbf{10.57} \\ \hline
& UniControl  & 64.50  & 64.49   & 64.48   & 64.48  & 64.48  & 64.55  & 64.54  & 64.54  & 64.53   & 64.51 \\
& Controlnet++    & 72.49     & -    & -     & -    & -    & -    & -    & -     & -      & -  \\
& OminiControl$^\dagger$    & \underline{82.25}    & \underline{79.1}   & \underline{78.27}   & 69.30   & 62.29   & \textbf{88.43} & \textbf{93.33} & \textbf{97.38} & \textbf{94.32} & \underline{82.74} \\
& PixelPonder$^\dagger$    & 79.98   & 78.61    & 75.49    & \underline{71.67}   & \underline{67.71}   & 80.65  & 85.53  & 89.89   & 91.75  & 80.14  \\
\multirow{-5}{*}{\rotatebox{90}{CLIP-I} \rotatebox{90}{$\xrightarrow{\hspace{0.65cm}}$}}  & \cellcolor{NatureGreen!30}\textbf{UniGen(ours)} & \cellcolor{NatureGreen!30}\textbf{84.06} & \cellcolor{NatureGreen!30}\textbf{81.01} & \cellcolor{NatureGreen!30}\textbf{79.96} & \cellcolor{NatureGreen!30}\textbf{73.23} & \cellcolor{NatureGreen!30}\textbf{69.52} & \cellcolor{NatureGreen!30}\underline{87.60}   & \cellcolor{NatureGreen!30}\underline{92.52} & \cellcolor{NatureGreen!30}\underline{96.88} & \cellcolor{NatureGreen!30}\underline{94.15}   & \cellcolor{NatureGreen!30}\textbf{84.33} \\ \hline
& UniControl   & 18.85   & 18.45   & 19.06   & 19.31    & 19.08   & 18.98   & 19.56  & 19.26   & 18.75  & 19.03 \\
& Controlnet++   & \underline{19.73}   & -   & -     & -    & -   & -    & -   & -   & -  & -  \\
& OminiControl$^\dagger$    & 18.65   & 18.25   & 19.21    & 19.22     & 19.35   & 19.06    & 19.18    & 18.91   & 19.10   & 18.99 \\
& PixelPonder$^\dagger$     & \textbf{20.0}  & \textbf{19.59} & \textbf{20.49} & \textbf{20.61} & \textbf{20.12} & \textbf{21.13} & \textbf{21.0}  & \textbf{20.53} & \textbf{21.05} & \textbf{20.50} \\
\multirow{-5}{*}{\rotatebox{90}{CLIP-T} \rotatebox{90}{$\xrightarrow{\hspace{0.65cm}}$}}  & \cellcolor{NatureGreen!30}\textbf{UniGen(ours)} & \cellcolor{NatureGreen!30}19.30  & \cellcolor{NatureGreen!30}\underline{18.70}   & \cellcolor{NatureGreen!30}\underline{19.80}   & \cellcolor{NatureGreen!30}\underline{19.99}    & \cellcolor{NatureGreen!30}\underline{19.76}  & \cellcolor{NatureGreen!30}\underline{19.66}   & \cellcolor{NatureGreen!30}\underline{19.58}   & \cellcolor{NatureGreen!30}\underline{19.53}   & \cellcolor{NatureGreen!30}\underline{19.28}  & \cellcolor{NatureGreen!30}\underline{19.51} \\ \hline
& UniControl  & 75.72   & 75.72   & 75.70   & 75.70   & 75.70    & 75.71   & 75.71   & 75.70    & 75.70   & 75.71\\
& Controlnet++   & 86.22   & -   & -   & -   & -   & -   & -   & -   & -  & - \\
& OminiControl$^\dagger$    & \underline{92.07}   & \underline{90.13}    & \underline{90.41}   & \underline{83.78}   & 75.10   & \textbf{93.38} & \textbf{97.47} & \textbf{98.20} & \textbf{98.57}  & \underline{91.01} \\
& PixelPonder$^\dagger$     & 89.33   & 88.07   & 86.06   & 81.44    & \underline{76.31}    & 88.24    & 93.5    & 95.54   & 97.06    &  88.39\\
\multirow{-5}{*}{\rotatebox{90}{DINO} \rotatebox{90}{$\xrightarrow{\hspace{0.5cm}}$}}    & \cellcolor{NatureGreen!30}\textbf{UniGen(ours)} & \cellcolor{NatureGreen!30}\textbf{92.50} & \cellcolor{NatureGreen!30}\textbf{90.66} & \cellcolor{NatureGreen!30}\textbf{90.72} & \cellcolor{NatureGreen!30}\textbf{85.44} & \cellcolor{NatureGreen!30}\textbf{81.11} & \cellcolor{NatureGreen!30}\underline{92.97}          & \cellcolor{NatureGreen!30}\underline{96.85}          & \cellcolor{NatureGreen!30}\underline{97.86}          & \cellcolor{NatureGreen!30}\underline{98.41}   &  \cellcolor{NatureGreen!30}\textbf{91.84}\\ \hline
\end{tabular}}
\caption{Comparison with state-of-the-art image-to-image conditional control models on the MultiGen-20M test set. $\dagger$ denotes the use of FLUX \cite{flux.1} as the base model.}
\label{tab:multigen}
\end{table*}

\begin{table*}[h!]
\centering
\begin{tabular}{ccccccccc}
\hline
Task                      & Method            & AES $\uparrow$ & IMG $\uparrow$ & CLIP Cap $\uparrow$ & L1 Raw $\downarrow$ & L1 Color $\uparrow$ & L1 Cond $\uparrow$ & Final Score $\uparrow$ \\ \hline
\multirow{7}{*}{Depth}    & ControlNet        & 5.34 & 51.52 & 0.26     & 0.16   & 0.83     & \textbf{0.93}    & 0.59        \\
& UniControl        & 5.19 & 48.82 & \textbf{0.27}     & 0.24   & {0.76}     & 0.92    & 0.58        \\
& Controlnet ++     & 5.30 & 50.48 & 0.26     & 0.20   & 0.80     & \textbf{0.93}    & 0.59        \\
& OminiControl$^\dagger$   & 5.42 & 51.51 & 0.25     & 0.15   & 0.86     & \textbf{0.93}    & 0.59        \\
& PixelPonder$^\dagger$    & 4.96 & 48.71 & 0.24     & 0.22   & 0.80     & {0.88}    & 0.56        \\
& \cellcolor{SoftBlue!50}\textbf{UniGen(ours)}      & \cellcolor{SoftBlue!50}5.34 & \cellcolor{SoftBlue!50}50.11 & \cellcolor{SoftBlue!50}0.26     & \cellcolor{SoftBlue!50}0.15   & \cellcolor{SoftBlue!50}0.85     & \cellcolor{SoftBlue!50}\textbf{0.93}    & \cellcolor{SoftBlue!50}0.59        \\
& \cellcolor{SoftBlue!50}\textbf{UniGen(ours)}$^\dagger$ & \cellcolor{SoftBlue!50}\textbf{5.48} & \cellcolor{SoftBlue!50}\textbf{52.78} & \cellcolor{SoftBlue!50}0.25     & \cellcolor{SoftBlue!50}\textbf{0.13}   & \cellcolor{SoftBlue!50}\textbf{0.87}     & \cellcolor{SoftBlue!50}\textbf{0.93}    & \cellcolor{SoftBlue!50}\textbf{0.64}        \\ \hline
\multirow{7}{*}{Canny}    & ControlNet        & 5.36 & 50.16 & \textbf{0.26}     & 0.16   & 0.84     & 0.93    & 0.59        \\
& UniControl        & 5.35 & 47.78 & \textbf{0.26}     & 0.22   & {0.79}     & 0.92    & 0.58        \\
& Controlnet ++     & 5.43 & 42.71 & \textbf{0.26}     & 0.20   & 0.80     & 0.91    & 0.57        \\
& OminiControl$^\dagger$  & 5.43 & 49.77 & 0.25     & 0.13   & 0.88     & \textbf{0.94}    & 0.59        \\
& PixelPonder$^\dagger$    & 4.85 & 49.25 & 0.24     & 0.22   & 0.81     & {0.90}    & 0.56        \\
& \cellcolor{SoftBlue!50}\textbf{UniGen(ours)}      & \cellcolor{SoftBlue!50}5.31 & \cellcolor{SoftBlue!50}49.74 & \cellcolor{SoftBlue!50}\textbf{0.26}     & \cellcolor{SoftBlue!50}0.14   & \cellcolor{SoftBlue!50}0.87     & \cellcolor{SoftBlue!50}\textbf{0.94}    & \cellcolor{SoftBlue!50}0.59        \\
& \cellcolor{SoftBlue!50}\textbf{UniGen(ours)}$^\dagger$ & \cellcolor{SoftBlue!50}\textbf{5.45} & \cellcolor{SoftBlue!50}\textbf{51.34} & \cellcolor{SoftBlue!50}0.25     & \cellcolor{SoftBlue!50}\textbf{0.11}   & \cellcolor{SoftBlue!50}\textbf{0.90}     & \cellcolor{SoftBlue!50}0.93    & \cellcolor{SoftBlue!50}\textbf{0.64}        \\ \hline
\multirow{7}{*}{Openpose} & ControlNet        & 5.37 & 49.93 & 0.26     & 0.21   & 0.78     & {0.97}    & 0.59        \\
& UniControl        & 5.12 & 47.42 & 0.27     & 0.29   & {0.71}     & {0.97}    & 0.56        \\
& Controlnet ++     & -    & -     & -        & -      & -        & -       & -           \\
& OminiControl$^\dagger$      & 5.39 & 48.11 & 0.26     & 0.21   & 0.79     & \textbf{0.98}    & 0.58        \\
& PixelPonder$^\dagger$       & 5.00 & 47.60 & \textbf{0.24}     & 0.23   & 0.80     & \textbf{0.98}    & 0.56        \\
& \cellcolor{SoftBlue!50}\textbf{UniGen(ours)}      & \cellcolor{SoftBlue!50}5.39 & \cellcolor{SoftBlue!50}48.28 & \cellcolor{SoftBlue!50}0.27     & \cellcolor{SoftBlue!50}0.22   & \cellcolor{SoftBlue!50}0.79     & \cellcolor{SoftBlue!50}\textbf{0.98}    & \cellcolor{SoftBlue!50}0.58        \\
& \cellcolor{SoftBlue!50}\textbf{UniGen(ours)}$^\dagger$ & \cellcolor{SoftBlue!50}\textbf{5.49} & \cellcolor{SoftBlue!50}\textbf{51.71} & \cellcolor{SoftBlue!50}0.26     & \cellcolor{SoftBlue!50}\textbf{0.20}   & \cellcolor{SoftBlue!50}\textbf{0.81}     & \cellcolor{SoftBlue!50}\textbf{0.98}    & \cellcolor{SoftBlue!50}\textbf{0.64}        \\ \hline
\end{tabular}
\caption{Performance evaluation of generation quality under three conditioning settings on the Subjects-200K dataset using the ICE-Bench evaluation protocol. $\dagger$ denotes the use of FLUX \cite{flux.1} as the base model.}
\label{tab:ice_subjects_res}
\end{table*}

\begin{table*}[h!]
\setlength{\tabcolsep}{5pt}
\centering
\begin{tabular}{ccccccccccc}
\hline
Task                      & Method            & MUSIQ $\uparrow$ & MAN-IQA $\uparrow$ & CLIP-IQA $\uparrow$ & LPIPS $\downarrow$ & LAION-AES $\uparrow$ & PSNR $\uparrow$  & NIMA $\uparrow$ & TOPIQ-NR $\uparrow$ & CLIPSCORE $\uparrow$ \\ \hline
\multirow{7}{*}{Depth}    & ControlNet        & 58.96 & 0.49   & 0.56     &   0.55    & 5.62      &   12.75    & 5.00 & 0.70           & 0.79       \\
& UniControl        & 59.98 & 0.48   & 0.52     & 0.63  & 5.74      & 10.28 & 4.87 & 0.71           & \textbf{0.81}       \\
& Controlnet ++     & \textbf{61.00} & 0.49   & 0.54     & 0.58  & 5.74      & 11.59 & 4.99 & 0.71           & 0.80       \\
& OminiControl      & 60.21 & 0.49   & 0.56     & \textbf{0.50}  & 5.62      & 14.16 & 4.89 & 0.71           &    0.77        \\
& PixelPonder       &  58.91     &  0.46      &   0.51       & 0.66      &  5.71         & 10.89      &  4.98    &   0.69             &    0.74        \\
& \cellcolor{SoftBlue!50}\textbf{UniGen(ours)}      & \cellcolor{SoftBlue!50}58.78 & \cellcolor{SoftBlue!50}0.49   & \cellcolor{SoftBlue!50}0.54     & \cellcolor{SoftBlue!50}0.52  & \cellcolor{SoftBlue!50}5.69      & \cellcolor{SoftBlue!50}13.45 & \cellcolor{SoftBlue!50}4.97 & \cellcolor{SoftBlue!50}0.71           &  \cellcolor{SoftBlue!50}  0.79  \\
& \cellcolor{SoftBlue!50}\textbf{UniGen(ours)}$^\dagger$ & \cellcolor{SoftBlue!50}60.66 & \cellcolor{SoftBlue!50}\textbf{0.52}   & \cellcolor{SoftBlue!50}\textbf{0.58}     & \cellcolor{SoftBlue!50}\textbf{0.50}  & \cellcolor{SoftBlue!50}\textbf{5.75}      & \cellcolor{SoftBlue!50}\textbf{14.57} & \cellcolor{SoftBlue!50}\textbf{5.04} & \cellcolor{SoftBlue!50}\textbf{0.73}           & \cellcolor{SoftBlue!50}0.78       \\ \hline
\multirow{7}{*}{Canny}    & ControlNet        & 58.16 & 0.48   & 0.55     &  0.51     & 5.63      &   13.32    & 4.98 & 0.70           & 0.79       \\
& UniControl        & 57.21 & 0.45   & 0.49     & 0.58  & 5.68      & 11.06 & 4.82 & 0.69           & \textbf{0.80}       \\
& Controlnet ++     & 53.38 & 0.39   & 0.45     & 0.60  & 5.66      & 11.43 & 4.81 & 0.68           & \textbf{0.80}       \\
& OminiControl      & 59.22 & 0.48   & 0.55     & 0.45  & 5.60      & 15.30 & 4.90 & 0.70           &    0.76        \\
& PixelPonder       &  \textbf{60.39}     &  0.48      &     0.52     &    0.61   & 5.63          & 11.13      & 5.00     &   0.70             &   0.74         \\
& \cellcolor{SoftBlue!50}\textbf{UniGen(ours)}      & \cellcolor{SoftBlue!50}58.76 & \cellcolor{SoftBlue!50}0.48   & \cellcolor{SoftBlue!50}0.54     & \cellcolor{SoftBlue!50}0.48  & \cellcolor{SoftBlue!50}5.67      & \cellcolor{SoftBlue!50}14.37 & \cellcolor{SoftBlue!50}4.99 & \cellcolor{SoftBlue!50}0.72           &   \cellcolor{SoftBlue!50}0.79     \\
& \cellcolor{SoftBlue!50}\textbf{UniGen(ours)}$^\dagger$ & \cellcolor{SoftBlue!50}59.89 & \cellcolor{SoftBlue!50}\textbf{0.52}   & \cellcolor{SoftBlue!50}\textbf{0.58}     & \cellcolor{SoftBlue!50}\textbf{0.42}  & \cellcolor{SoftBlue!50}\textbf{5.71}      & \cellcolor{SoftBlue!50}\textbf{16.19} & \cellcolor{SoftBlue!50}\textbf{5.03} & \cellcolor{SoftBlue!50}\textbf{0.73}           & \cellcolor{SoftBlue!50}0.78       \\ \hline
\multirow{7}{*}{Openpose} & ControlNet        & 57.49 & 0.47   & 0.55     &  0.66     & 5.58      &  10.75     & 5.00 & 0.69           & 0.80 \\
& UniControl        & 59.38 & 0.45   & 0.51     & 0.75  & 5.76      & 8.56  & 4.76 & 0.71           & \textbf{0.82} \\
& Controlnet ++     & -     & -      & -        & -     & -         & -     & -    & -              & -          \\
& OminiControl      & 57.58 & 0.44   & 0.53     & 0.68  & 5.57      & 11.20 & 4.72 & 0.68           &    0.79        \\
& PixelPonder       &   57.62    &  0.45      &  0.50        & 0.69      & 5.70          &  10.41     &  \textbf{4.95}    &   0.68             & 0.75           \\
& \cellcolor{SoftBlue!50}\textbf{UniGen(ours)}      & \cellcolor{SoftBlue!50}56.78 & \cellcolor{SoftBlue!50}0.47   & \cellcolor{SoftBlue!50}0.54     & \cellcolor{SoftBlue!50}0.65  & \cellcolor{SoftBlue!50}5.70      & \cellcolor{SoftBlue!50}10.86 & \cellcolor{SoftBlue!50}\textbf{4.95} & \cellcolor{SoftBlue!50}0.70           &     \cellcolor{SoftBlue!50}0.81       \\
& \cellcolor{SoftBlue!50}\textbf{UniGen(ours)}$^\dagger$ & \cellcolor{SoftBlue!50}\textbf{59.83} & \cellcolor{SoftBlue!50}\textbf{0.52}   & \cellcolor{SoftBlue!50}\textbf{0.57}     & \cellcolor{SoftBlue!50}\textbf{0.64}  & \cellcolor{SoftBlue!50}\textbf{5.84}      & \cellcolor{SoftBlue!50}\textbf{11.69} & \cellcolor{SoftBlue!50}4.92 & \cellcolor{SoftBlue!50}\textbf{0.72}           & \cellcolor{SoftBlue!50}0.79       \\ \hline
\end{tabular}
\caption{Performance evaluation of generation quality under three conditioning settings on the Subjects-200K dataset using the IQA-Pytorch assessment framework. $\dagger$ denotes the use of FLUX \cite{flux.1} as the base model.}
\label{tab:iqa_subjects_res}
\end{table*}

\begin{table}[h!]
\centering
{\begin{tabular}{cccc}
\hline
\begin{tabular}[c]{@{}c@{}}Condition\\ Nums\end{tabular}   & Method   & Params(B) $\downarrow$ & Inference Times $\downarrow$ \\ \hline
\multirow{3}{*}{3}  & ControlNet   & 6.03      & 58.12           \\
                    & UniControl & 4.8     & 13.30           \\
                    & OminiControl$^\dagger$ & 11.93     & 13.06           \\
& \cellcolor{SoftBlue!60}\textbf{UniGen}       & \cellcolor{SoftBlue!60}\textbf{3.48}  & \cellcolor{SoftBlue!60}\textbf{6.82}    \\ \hline
\multirow{3}{*}{12} & ControlNet   & 17.38     & 59.16           \\
                    & UniControl & 4.94     & 14.86          \\
                    & OminiControl$^\dagger$ & 12.07     & 15.74           \\
& \cellcolor{SoftBlue!60}\textbf{UniGen}       & \cellcolor{SoftBlue!60}\textbf{4.07}  & \cellcolor{SoftBlue!60}\textbf{13.96}      \\ \hline
\end{tabular}}
\caption{Complexity analysis between ControlNet, UniControl, OminiControl and UniGen. $\dagger$ denotes the use of FLUX \cite{flux.1} as the base model.}
\label{tab:params_inference}
\end{table}

\begin{table*}[h!]
\setlength{\tabcolsep}{4.5pt}
\centering
\begin{tabular}{ccccccccccc}
\hline
\multicolumn{11}{c}{\slashrowcellfixed{Evaluate on traditional metrics}} \\ \hline
Task                             & Method     & Base Model & PNSR $\uparrow$  & SSIM $\uparrow$ & FID $\downarrow$  & CLIP-I $\uparrow$ & CLIP-T $\uparrow$ & LPIPS $\downarrow$ & DINO $\uparrow$  & Param (B) $\downarrow$          \\ \hline
\multirow{2}{*}{Depth + Canny}   & UniCombine \cite{Unicombine} & Flux .1 \cite{flux.1}    & 13.66 & 0.45  & 11.09    & 92.02    & 29.90  & 0.56     & \textbf{97.03}          &   11.93            \\
& \cellcolor{SoftBlue!50}  UniGen(Ours)   & \cellcolor{SoftBlue!50}SD 3.5 \cite{sd35}   & \cellcolor{SoftBlue!50}\textbf{15.65} & \cellcolor{SoftBlue!50}\textbf{0.62}  & \cellcolor{SoftBlue!50}\textbf{9.03}     & \cellcolor{SoftBlue!50}92.73    & \cellcolor{SoftBlue!50}\textbf{30.45}  & \cellcolor{SoftBlue!50}\textbf{0.43}     & \cellcolor{SoftBlue!50}96.54     &  \cellcolor{SoftBlue!50}\textbf{3.41}     \\ \hline
\multirow{2}{*}{Subject + Depth} & UniCombine \cite{Unicombine} & Flux .1 \cite{flux.1}    & 13.18 & 0.42  & \textbf{11.56}    & \textbf{91.25}    & 29.89  & 0.58     & \textbf{96.27}          &  11.93          \\
& \cellcolor{SoftBlue!50} UniGen(Ours)   & \cellcolor{SoftBlue!50}SD 3.5 \cite{sd35} & \cellcolor{SoftBlue!50}\textbf{13.28} & \cellcolor{SoftBlue!50}\textbf{0.48}  & \cellcolor{SoftBlue!50}11.69    & \cellcolor{SoftBlue!50}89.36    & \cellcolor{SoftBlue!50}\textbf{30.72}  & \cellcolor{SoftBlue!50}\textbf{0.52}     & \cellcolor{SoftBlue!50}94.28    & \cellcolor{SoftBlue!50}\textbf{3.41}    \\ \hline
\multirow{2}{*}{Subject + Canny} & UniCombine \cite{Unicombine} & Flux .1 \cite{flux.1}   & 13.66 & 0.45  & \textbf{9.51}    & \textbf{92.52}    & 29.71  & 0.58     & \textbf{96.63}          &   11.93 \\
& \cellcolor{SoftBlue!50} UniGen(Ours)   & \cellcolor{SoftBlue!50}SD 3.5 \cite{sd35}  & \cellcolor{SoftBlue!50}\textbf{14.59} & \cellcolor{SoftBlue!50}\textbf{0.57}  & \cellcolor{SoftBlue!50}9.66     & \cellcolor{SoftBlue!50}91.86    & \cellcolor{SoftBlue!50}\textbf{30.58}  & \cellcolor{SoftBlue!50}\textbf{0.47}     & \cellcolor{SoftBlue!50}95.27     &   \cellcolor{SoftBlue!50}\textbf{3.41}   \\ \hline
\multicolumn{11}{c}{\slashrowcellfixed{Evaluate on ICE Bench \cite{ice_bench} (ICCV 2025)}} \\ \hline
Task                             & Method     & Base Model    & AES $\uparrow$  & IMG $\uparrow$  & CLIP Cap $\uparrow$ & CLIP Src $\uparrow$ & L1 Raw $\downarrow$ & L1 Color $\uparrow$ & Final Score $\uparrow$ & Param (B) $\downarrow$ \\ \hline
\multirow{2}{*}{Depth + Canny}   & UniCombine \cite{Unicombine} & Flux .1 \cite{flux.1}    & 5.37  & \textbf{58.16} & 0.25     & - & 0.14   & {0.86}     & \textbf{0.60} & 11.93            \\
& \cellcolor{SoftBlue!50} UniGen(Ours)   & \cellcolor{SoftBlue!50}SD 3.5 \cite{sd35}  & \cellcolor{SoftBlue!50}\textbf{5.46}  & \cellcolor{SoftBlue!50}48.84 & \cellcolor{SoftBlue!50}\textbf{0.26}     &  \cellcolor{SoftBlue!50}-& \cellcolor{SoftBlue!50}\textbf{0.12}   & \cellcolor{SoftBlue!50}\textbf{0.88}     & \cellcolor{SoftBlue!50}0.59 & \cellcolor{SoftBlue!50}\textbf{3.41}            \\ \hline
\multirow{2}{*}{Subject + Depth} & UniCombine \cite{Unicombine} & Flux .1 \cite{flux.1}   & 5.04  & \textbf{52.76} & 0.25     & \textbf{0.72} & \textbf{0.15}   & \textbf{0.84}     & \textbf{0.57} & 11.93            \\
& \cellcolor{SoftBlue!50} UniGen(Ours)   & \cellcolor{SoftBlue!50}SD 3.5 \cite{sd35} & \cellcolor{SoftBlue!50}\textbf{5.32}  & \cellcolor{SoftBlue!50}48.09 & \cellcolor{SoftBlue!50}\textbf{0.26}     &  \cellcolor{SoftBlue!50}0.68  & \cellcolor{SoftBlue!50}0.16   & \cellcolor{SoftBlue!50}\textbf{0.84}     & \cellcolor{SoftBlue!50}\textbf{0.57} & \cellcolor{SoftBlue!50}\textbf{3.41}  \\ \hline
\multirow{2}{*}{Subject + Canny} & UniCombine \cite{Unicombine} & Flux .1 \cite{flux.1}  & 5.09  & \textbf{55.52} & 0.25     &  \textbf{0.72}   & 0.14   & {0.85}     & \textbf{0.58} & 11.93            \\
& \cellcolor{SoftBlue!50} UniGen(Ours)   & \cellcolor{SoftBlue!50}SD 3.5 \cite{sd35} & \cellcolor{SoftBlue!50}\textbf{5.36}  & \cellcolor{SoftBlue!50}48.76 &\cellcolor{SoftBlue!50}\textbf{0.26}     &\cellcolor{SoftBlue!50}0.70  & \cellcolor{SoftBlue!50}\textbf{0.13}   & \cellcolor{SoftBlue!50}\textbf{0.87}     & \cellcolor{SoftBlue!50}\textbf{0.58}  & \cellcolor{SoftBlue!50}\textbf{3.41}            \\ \hline
\end{tabular}
\caption{Performance evaluation of multi-condition generation results using both traditional metrics and the ICE-Bench evaluation protocol.}
\label{tab:multi_condition}
\end{table*}

\subsection{Comparison and Analysis with state-of-the-art Methods}
\subsubsection{Comparison and Analysis of Single-Condition Controllable Image Generation Performance}
\label{sec:compare_sota_single_condition}
We train and evaluate our method on Subjects-200K \cite{ominicn} and MultiGen-20M \cite{cn++}. For fairness, we remove all test samples and randomly sample 160K instruction–target–condition image triplets per condition from the remaining data for training. We compare against traditional baselines like UniControl \cite{Unicontrol} and ControlNet++ \cite{cn++}, as well as recent state-of-the-art methods such as OminiControl \cite{ominicn} and PixelPonder \cite{Pixelponder}.

First, as shown in Table \ref{tab:ss200k}, we conducted a comparative evaluation on the Subjects-200K and MultiGen-20M datasets for three types of conditions (Depth, Canny, and OpenPose) using traditional metrics. It is clear and intuitive that our method outperforms existing approaches across all three condition types. Specifically, when combined with the current best base model, Flux .1 \cite{flux.1}, our method demonstrates a significant advantage, achieving the highest average performance across all metrics and substantially surpassing other methods based on the same base model. Meanwhile, when applied to the smaller Stable Diffusion 3.5 model, our method still achieves competitive, near-optimal performance. This indicates that our approach not only exhibits strong generalization capability but also effectively enables controllable image generation according to different condition types.

Next, as shown in Table \ref{tab:multigen}, we compare image generation quality across the remaining nine condition types on the MultiGen-20M dataset. Even using the smaller Stable Diffusion as the base model, our method achieves the best or second-best performance in multiple evaluation metrics across these nine condition types, in some cases significantly outperforming methods based on Flux .1. Overall, our method ensures precise condition control and high-quality image content while maintaining strong generalization across diverse condition types.

Considering that traditional metrics may be imperfect in evaluating image quality and aesthetics, we further conducted comprehensive performance assessments under the ICE-Bench \cite{ice_bench} and IQA-Torch frameworks. As shown in Table \ref{tab:ice_subjects_res}, ICE-Bench provides multi-faceted evaluation, including aesthetics, image quality, instruction adherence, and condition control. Table \ref{tab:ice_subjects_res} clearly shows that our method achieves the best or second-best performance under two different base models. Notably, in terms of aesthetics and image quality, our method significantly outperforms the next-best approach. For example, under Depth conditions, compared to OminiControl with the same base model, our method improves image quality by +1.27; under Canny and OpenPose conditions, improvements of +1.5 and +3.6 are observed, respectively.

Similarly, as shown in Table \ref{tab:iqa_subjects_res}, evaluations conducted with IQA-PyTorch-based image quality metrics indicate that our method achieves the best performance across multiple dimensions, including aesthetics, image quality, and perceptual distance, substantially surpassing related methods based on similar base models.
However, in prior evaluations, we noticed that our method is relatively less competitive in metrics related to adherence to textual instructions, generally ranking second rather than first. We attribute this to the iterative interaction mechanism in our WeaveNet architecture for condition feature fusion and injection. Compared to traditional ControlNet \cite{controlnet} or LoRA-based \cite{lora,ominicn} methods, WeaveNet focuses more on processing visual condition information during intermediate steps, which reduces the influence of the original textual instructions and decreases reliance on prompts during condition control. Therefore, compared with methods that employ independent branches for prompt-based global generation, our method may perform slightly lower in aligning image content with text instructions, leading to suboptimal CLIP-T and CLIPScore results. This issue is further discussed in Section \ref{sec:limitation}.

Finally, we analyze model parameters and inference time. As shown in Table \ref{tab:params_inference}, the traditional ControlNet \cite{controlnet} architecture introduces significant parameter redundancy, and inference time increases with the number of condition types. This is primarily because ControlNet relies on parallel computation with relatively independent processing branches; after independently extracting condition features via ControlNet branches, the features are interactively injected into the base model, which inevitably increases inference time. OminiControl, based on the Flux model, trains separate LoRA modules \cite{lora} for each condition type. While this introduces fewer additional parameters, aligning with LoRA's design goals, it is limited in feature processing and overall performance. In contrast, our method, even when applied to a smaller base model with fewer parameters and lower intrinsic capability than Flux, achieves superior performance across multiple evaluation metrics and condition types.

Although our design introduces more parameters compared to LoRA \cite{lora}, we only require a single independent ControlNet architecture with additional expert control units as the number of conditions increases. This keeps the parameter growth minimal while maintaining comparable parameter increments to LoRA. Importantly, our approach achieves significantly greater performance gains compared to LoRA-based methods. As shown in Table \ref{tab:multigen}, when scaling to 12 tasks, our model introduces an additional 0.59B parameters, while OminiControl adds 0.14B. Although our parameter increase is four times larger, our method is built on a much smaller 2.5B-parameter Stable Diffusion backbone yet achieves higher performance than OminiControl, which is based on the larger Flux backbone. For example, in terms of FID, our model outperforms OminiControl by 4.37, 7.66, 6.92, 19.32, and 50.84 on the Hed, HedSketch, Normal, Seg, Bbox, and Outpainting tasks, respectively. Note that these tasks are more complex and challenging than Depth, Canny, and OpenPose, making them a stronger indicator of the model’s generative quality and controllability.

\subsubsection{Comparison and Analysis of Multi-Condition Controllable Image Generation Performance}
\label{sec:compare_sota_multi_condition}
To provide a more comprehensive evaluation of our approach, we further extend it to multi-condition joint control tasks. Compared with single-condition control, multi-condition joint control introduces more constraints and imposes stronger requirements on spatial alignment and subject consistency. Although one might expect that adding more conditions would simplify generation and improve performance, the opposite is true. Multi-condition joint control strictly restricts the model’s output—for example, enforcing consistency across subjects, spatial depth, and texture boundaries. As a result, many image generation methods that perform well under single-condition control fail to produce valid outputs in this setting.

Since this is an emerging task, there are few comparable reference methods. Therefore, we follow the evaluation setup of the UniCombine \cite{Unicombine} and conduct assessments under two different evaluation protocols. Considering that traditional approaches lose controllability in this task and cannot produce usable results, we do not include them in our comparisons. Instead, we focus on evaluating against UniCombine, the most recent and competitive method. Note that UniCombine, similar to OminiControl, is built on the larger and more capable Flux.1 base model, and achieves multi-condition joint control by independently training multiple LoRA components.

As shown in Table \ref{tab:multi_condition}, under the standard evaluation metrics, our method consistently achieves better performance across multiple metrics in three multi-condition control tasks. It should be noted that our model uses substantially fewer parameters than UniCombine, reducing the parameter count by 8.52B. Despite this reduction, the performance gap on several metrics remains minimal. Even in cases where our method is slightly weaker, the difference is within a margin of 1\%.

We further conduct a comprehensive evaluation using ICE-Bench. The results show that our approach achieves advantages in several aspects, including aesthetics and conditional controllability. Notably, our method demonstrates a significant improvement in aesthetic scores across all three multi-condition control tasks compared to UniCombine. As discussed in Section \ref{sec:qualitative_ana_multi_img}, multi-condition tasks impose stronger constraints, making our model produce more natural outputs. In contrast, UniCombine tends to overfit the conditional information, which leads to reduced aesthetic quality.

In general, we conducted comprehensive evaluations across multiple datasets, tasks, and metric systems. Our method consistently achieves significant improvements and, in many cases, even exceeds approaches with larger parameter counts and stronger base-model capabilities. This demonstrates that our method can effectively follow conditioning instructions while generating content that better aligns with aesthetic evaluation standards. Moreover, when extending from a single condition to multiple conditions, our approach introduces only a small increase in parameters—comparable to LoRA-based methods—yet delivers substantially greater performance gains.

\subsection{Ablation Studies}
To comprehensively validate the effectiveness of each module, we conducted an extensive ablation study from five perspectives: the impact of the WeaveNet architecture, the roles of the Modulated Expert and Rotary Position Embedding (RoPE), the necessity of the Shared Expert, and the influence of the number of experts and conditional control layers.

\subsubsection{Ablation Analysis of the WeaveNet Architecture} In traditional ControlNet, conditional feature processing is performed in parallel with the backbone network. Specifically, an independent copy of the backbone is used as a conditional control unit to extract features, which are then added to the features of each corresponding layer of the backbone to achieve global conditional feature injection. First, we evaluate using only the ControlNet architecture, without incorporating our proposed CoMoE module. As shown in the upper part of Table \ref{tab:compare_cn}, without CoMoE, simply maintaining the parallel design of ControlNet yields inferior performance compared to our method. For example, in terms of FID, our method outperforms the traditional ControlNet across Depth, Canny, and Openpose by 1.34, 0.82, and 0.68, respectively.

To further validate the advantage of the WeaveNet architecture, we also modified WeaveNet to use the traditional ControlNet-style feature injection, keeping the backbone network and conditional feature network independent. As shown in the lower part of Table \ref{tab:compare_cn}, even under the WeaveNet framework, switching to the parallel feature injection of ControlNet still produces a significant performance gain. This demonstrates that, compared to traditional ControlNet, WeaveNet better integrates the backbone and conditional control networks, enabling more precise and effective conditional control. In contrast, traditional ControlNet tends to process the two branches independently, which limits both feature fusion and accurate conditional control.

\begin{table}[h!]
\setlength{\tabcolsep}{0pt} 
\centering
% \begin{tabular}{ccccccc}
\begin{tabular}{*{7}{>{\centering\arraybackslash}p{1.27cm}}} 
\hline
\multicolumn{7}{c}{\gridcell{Wo. CoMoE}} \\ \hline
Con.  & Me.  & SSIM $\uparrow$  & FID $\downarrow$ & CLIP-I $\uparrow$  & CLIP-T $\uparrow$      & DINO $\uparrow$ \\ \hline
\multirow{2}{*}{Depth}    & CN & \textbf{0.49}    & 12.6   & 88.72          & \textbf{20.81}     & 93.9    \\
& \cellcolor{LightGray!80}WN   & \cellcolor{LightGray!80}\textbf{0.49}    & \cellcolor{LightGray!80}\textbf{11.26}  & \cellcolor{LightGray!80}\textbf{89.14}   & \cellcolor{LightGray!80}19.51     & \cellcolor{LightGray!80}\textbf{94.31}   \\ \hline
\multirow{2}{*}{Canny}    & CN & \textbf{0.55}    & 10.34  & 91.31          & \textbf{21.27}     & 94.88   \\
& \cellcolor{LightGray!80}WN   & \cellcolor{LightGray!80}\textbf{0.55}    & \cellcolor{LightGray!80}\textbf{9.52}   & \cellcolor{LightGray!80}\textbf{91.61} & \cellcolor{LightGray!80}20.21     & \cellcolor{LightGray!80}\textbf{95.20}    \\ \hline
\multirow{2}{*}{Openpose} & CN & 0.37    & 16.36  & 82.21          & \textbf{20.93}     & 87.54   \\
& \cellcolor{LightGray!80}WN   & \cellcolor{LightGray!80}\textbf{0.38}  & \cellcolor{LightGray!80}\textbf{15.68}  & \cellcolor{LightGray!80}\textbf{82.72}          & \cellcolor{LightGray!80}20.55     & \cellcolor{LightGray!80}\textbf{88.11}   \\ \hline
\multicolumn{7}{c}{\gridcell{W. CoMoE}} \\ \hline
Con.  & Me.  & SSIM $\uparrow$  & FID $\downarrow$ & CLIP-I $\uparrow$  & CLIP-T $\uparrow$      & DINO $\uparrow$ \\ \hline
\multirow{2}{*}{Depth}    & CN & 0.47    & 12.29   & 88.42          & \textbf{20.41}     & 93.51    \\
& \cellcolor{LightGray!80}WN   & \cellcolor{LightGray!80}\textbf{0.49}    & \cellcolor{LightGray!80}\textbf{11.26}  & \cellcolor{LightGray!80}\textbf{89.14}   & \cellcolor{LightGray!80}19.51     & \cellcolor{LightGray!80}\textbf{94.31}   \\ \hline
\multirow{2}{*}{Canny}    & CN & 0.53    & 10.48  & 91.14          & \textbf{20.41}     & 94.72   \\
& \cellcolor{LightGray!80}WN   & \cellcolor{LightGray!80}\textbf{0.55}    & \cellcolor{LightGray!80}\textbf{9.52}   & \cellcolor{LightGray!80}\textbf{91.61} & \cellcolor{LightGray!80}20.21     & \cellcolor{LightGray!80}\textbf{95.20}    \\ \hline
\multirow{2}{*}{Openpose} & CN & 0.36    & 16.27  & 82.10          & 20.42     & 87.58   \\
& \cellcolor{LightGray!80}WN   & \cellcolor{LightGray!80}\textbf{0.38}  & \cellcolor{LightGray!80}\textbf{15.68}  & \cellcolor{LightGray!80}\textbf{82.72}          & \cellcolor{LightGray!80}\textbf{20.55}     & \cellcolor{LightGray!80}\textbf{88.11}   \\ \hline
\end{tabular}
\caption{Comparison of ControlNet and WeaveNet under different conditional settings on the Subjects-200K test set.}
\label{tab:compare_cn}
\end{table}

\subsubsection{Ablation Analysis of Modulated Experts Module} 
\label{sec:ablation_moe_rope}
As shown in the first and second rows of Table \ref{tab:ablation_modulated_rope}, introducing the Modulated Expert does not always result in significant performance improvements. This is because the Modulated Expert is designed to handle multiple types of conditions simultaneously, mitigating the resource overhead caused by replicating the backbone or creating multiple preprocessing modules for different condition types. Its main role is to enhance model compatibility.

However, we observe that adding RoPE (Rotary Position Embedding) leads to a significant performance improvement. We attribute this to the feature alignment process in the Modulated Expert. During alignment, visual representations are tokenized and processed independently, which inevitably leads to the loss of spatial information. When features are later aggregated to reconstruct the original representation, this loss can misalign or weaken the spatial constraints provided by conditional information. Introducing RoPE before the Modulated Expert preserves spatial position information and strengthens spatial constraints of the conditional input, resulting in the significant improvement seen in the third row of Table \ref{tab:ablation_modulated_rope}.

\begin{table}[h!]
\centering
\begin{tabular}{ccccccc}
\hline
Module & SSIM $\uparrow$  & FID $\downarrow$ & CLIP-I $\uparrow$  & CLIP-T $\uparrow$      & DINO $\uparrow$ \\ \hline
-    & 0.48          & 11.82  & 88.83          & \textbf{20.63}          & 94.09   \\
w. MoE    & \textbf{0.50} & 11.71  & 88.96          & 20.60 & 94.08   \\

\cellcolor{LightGray}w. MoE/RoPE     & \cellcolor{LightGray}0.49          & \cellcolor{LightGray}\textbf{11.26}  & \cellcolor{LightGray}\textbf{89.14} & \cellcolor{LightGray}19.51         & \cellcolor{LightGray}\textbf{94.31}   \\ \hline
\end{tabular}
\caption{Ablation study on the effectiveness of the Modulated Expert (MoE) and Rotational Position Encoding (RoPE).}
\label{tab:ablation_modulated_rope}
\end{table}

\subsubsection{Ablation Analysis of Shared Expert Module} We note that the Modulated Expert module, when processing conditional visual representations, tends to first scatter these representations and then re-aggregate them by token for handling similar types of visual tokens. However, this approach inevitably raises concerns about disrupting the global structure of the conditional visual tokens. As demonstrated earlier, scattering the entire set of visual tokens for preprocessing similar token types can lead to a loss of spatial information. To address this, we introduce a Shared Expert module. As shown in Table \ref{tab:ablation_shared_expert}, the evaluation metrics of generated images improve significantly after incorporating the Shared Expert. This improvement is primarily because the Shared Expert processes the conditional visual representations from a holistic perspective and serves as a complement to the Modulated Expert module. Consequently, it effectively mitigates the loss and confusion of global spatial constraints caused by token scattering.

\begin{table}[h!]
\centering
\begin{tabular}{cccccc}
\hline
\begin{tabular}[c]{@{}c@{}}Shared \\ Expert\end{tabular} & SSIM $\uparrow$  & FID $\downarrow$ & CLIP-I $\uparrow$  & CLIP-T $\uparrow$      & DINO $\uparrow$ \\ \hline
\ding{55}         & 0.49    & 12.31  & 88.75          & \textbf{20.14}     & 93.70   \\

\cellcolor{LightGray}\ding{51}         & \cellcolor{LightGray}0.49    & \cellcolor{LightGray}\textbf{11.26}  & \cellcolor{LightGray}\textbf{89.14} & \cellcolor{LightGray}19.51     & \cellcolor{LightGray}\textbf{94.31}   \\ \hline
\end{tabular}
\caption{Ablation study on the effectiveness of the Shared Expert.}
\label{tab:ablation_shared_expert}
\end{table}

\subsubsection{Ablation Analysis of the Number of Experts and Conditional Control Layers}
Finally, we conducted a comprehensive ablation study to analyze the impact of the number of experts in the Modulated Expert module and the number of conditional control units in the WeaveNet architecture. As shown in Table \ref{tab:ablation_moe}, using six experts leads to the highest image generation quality. Increasing or decreasing the number of experts results in notable performance degradation. We attribute this to two factors: (1) Excessive experts overly fragment the visual tokens, weakening inter-token consistency and causing the conditional visual representations to lose coherence and spatial information. (2) Too few experts reduce the diversity among conditional tokens, making it difficult for the model to distinguish foreground from background, ultimately harming generation quality under conditional constraints.

\begin{figure}[h!]
    \centering
    \includegraphics[width=\linewidth]{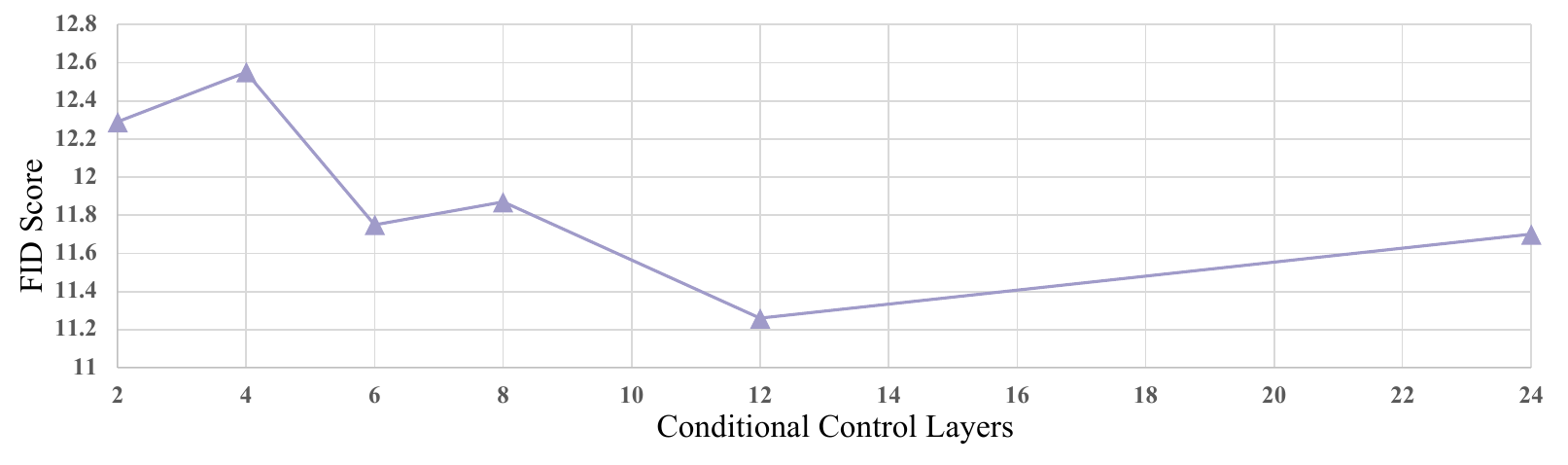}
    \caption{Ablation study on the number of conditional control layers in WeaveNet.}
    \label{fig:ablation_layers}
\end{figure}

Similarly, for the number of conditional control units, Fig. \ref{fig:ablation_layers} shows that using 12 layers achieves the best performance, representing a balanced trade-off between accuracy and computational cost. Fully replicating all base-model blocks for conditional feature extraction and fusion introduces a risk of overfitting to the conditioning signals, which can degrade the model’s inherent generation quality. Consequently, as the number of control layers increases beyond this point, we observe a slight drop in performance.

\begin{table}[h!]
\centering
\begin{tabular}{cccc}
\hline
Expert Numbers & SSIM $\uparrow$  & FID $\downarrow$   & DINO $\uparrow$        \\ \hline
2           & 0.49          & 11.80        & 94.28          \\
3           & \textbf{0.50} & 11.90        & 94.17          \\
4           & 0.49          & 11.72        & 94.25          \\
\cellcolor{LightGray}6           & \cellcolor{LightGray}0.49          & \cellcolor{LightGray}\textbf{11.26}   & \cellcolor{LightGray}\textbf{94.31} \\
8           & 0.49          & 11.74        & 94.20          \\
9           & 0.48          & 12.02        & 94.01          \\
10          & 0.48          & 11.66        & 94.11          \\
12          & 0.48          & 11.72        & 94.20          \\ \hline
\end{tabular}
\caption{Ablation study on the number of Modulated Expert modules.}
\label{tab:ablation_moe}
\end{table}

\begin{figure}[h!]
    \centering
    \includegraphics[width=\linewidth]{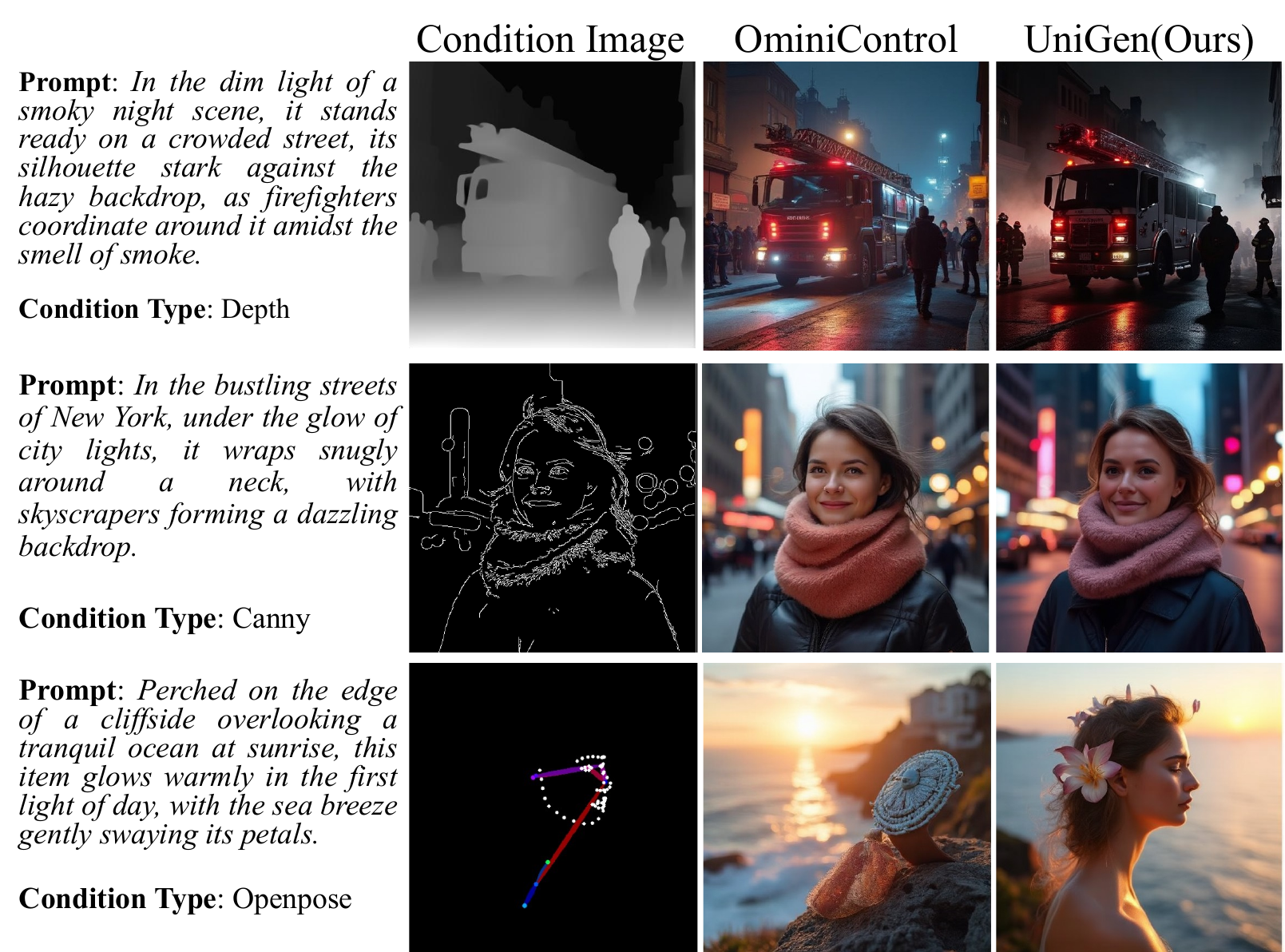}
    \caption{Comparison of generated results under single-condition control.}
    \label{fig:vis_compare}
\end{figure}

\subsection{Qualitative Analysis}
\subsubsection{Qualitative Analysis of Single-Condition Controlable Image Generation}
\label{sec:qualitative_ana_single_img}
In single-condition controlled generation tasks, condition types are more diverse and often emphasize spatial constraints. As a result, different conditions lead to different focal points and levels of controllability. As shown in Fig. \ref{fig:vis_compare}, conditions such as Depth and Canny impose stronger spatial constraints (e.g., the first and second rows), requiring the model to align more precisely with the spatial-level guidance provided by the conditions. This imposes higher demands on the model’s spatial understanding and alignment capabilities.
We observe that under Depth and Canny, the strong spatial constraints lead all models to produce images with largely consistent global layouts. However, our model better captures the holistic scene structure and semantic context.
In contrast, OpenPose provides weaker spatial constraints and focuses mainly on human pose and coarse spatial positioning, resulting in more limited conditional information. As shown in the third row, our method follows the spatial instructions more faithfully than the OminiControl and achieves better alignment even with sparse spatial constraints, enabling high-quality controllable image generation.

Additional comparisons with various models, including Hunyuan-DiT \cite{hunyuan} and UniControl \cite{Unicontrol}, are provided in the Appendix. We also present results under nine other conditional control settings for comprehensive evaluation.

\subsubsection{Qualitative Analysis of Multi-Condition Controlable Image Generation}
\label{sec:qualitative_ana_multi_img}
Compared with single-condition image generation, multi-condition generation provides more direct and diverse forms of control. For example, in tasks combining Subject \& Depth or Subject \& Canny, Depth and Canny offer spatial constraints, while Subject specifies the semantic identity of key objects. This setting imposes new and higher requirements on both subject consistency and adherence to conditional instructions.
Among these conditions, Depth \& Canny provide stronger spatial constraints and are therefore easier for the model to satisfy compared with the Subject condition in multi-condition tasks.

\begin{figure}[h!]
    \centering
    \includegraphics[width=\linewidth]{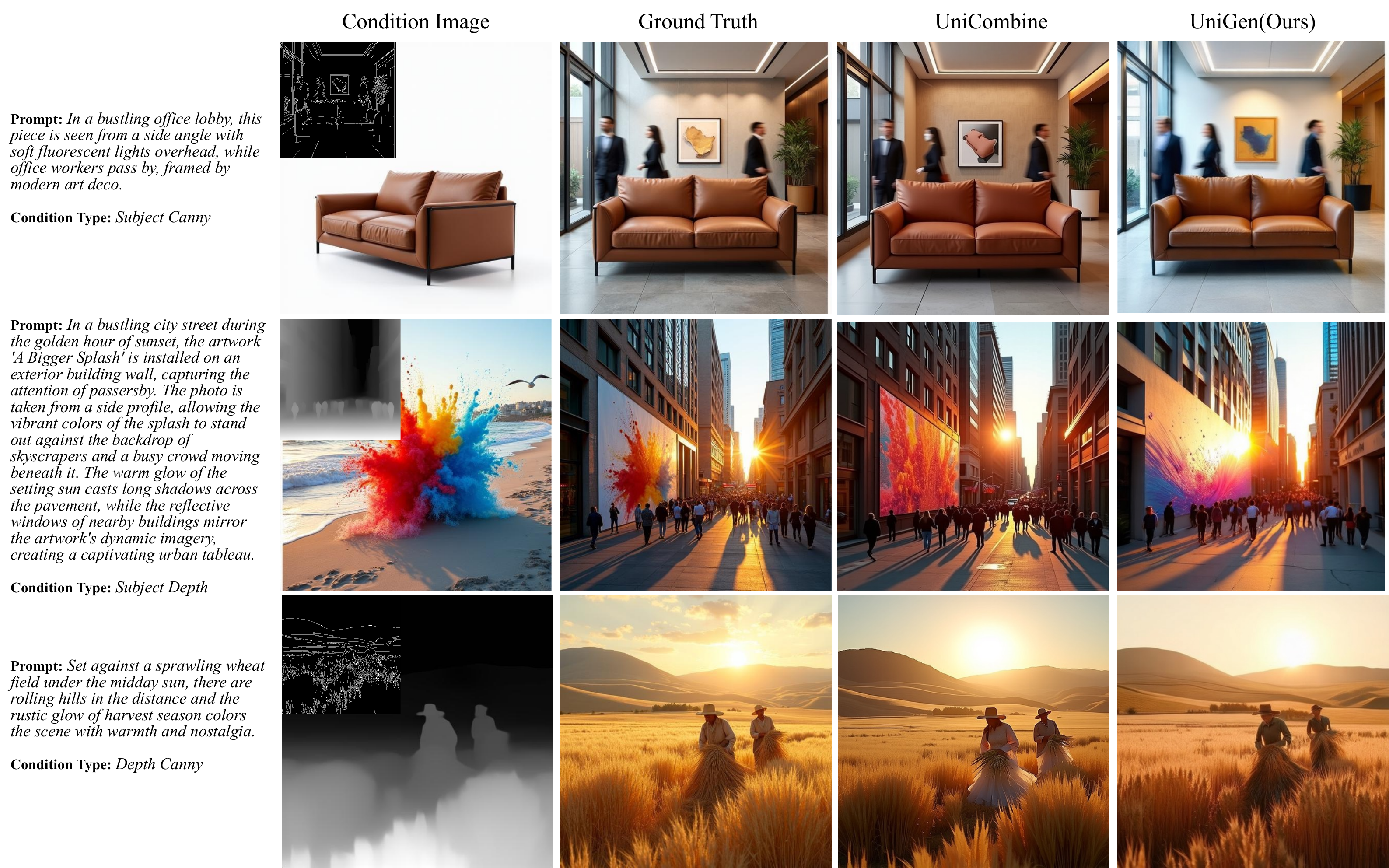}
    \caption{Comparison of generated results under multi-condition control.}
    \label{fig:vis_multicondition}
\end{figure}

As shown in Fig. \ref{fig:vis_multicondition}, we present model outputs under three types of multi-condition settings. In the Subject–Canny case, when Canny delivers strong spatial structure, the generated images accurately align with the provided Subject information, achieving consistent and controllable subject appearance. However, when the spatial constraint is weak, such as in the second row of Fig. \ref{fig:vis_multicondition} where Depth only regulates global scene layout and the Subject lies in a region without explicit spatial structure (e.g., a wall), our method preserves the subject information more effectively than UniCombine.
Moreover, for relatively simple constraints such as Depth \& Canny alone (third row of Fig. \ref{fig:vis_multicondition}), although both our method and UniCombine can fit the spatial conditions well, our approach generates objects with higher structural integrity and realism, resulting in outputs that more faithfully match the scene context and textual prompts.

\subsection{Failure Case Analysis}
\label{sec:error_sample_analysis}

\subsubsection{Failure Case Analysis of Single-Condition Controlable Image Generation}
\begin{figure}[h!]
    \centering
    \includegraphics[width=\linewidth]{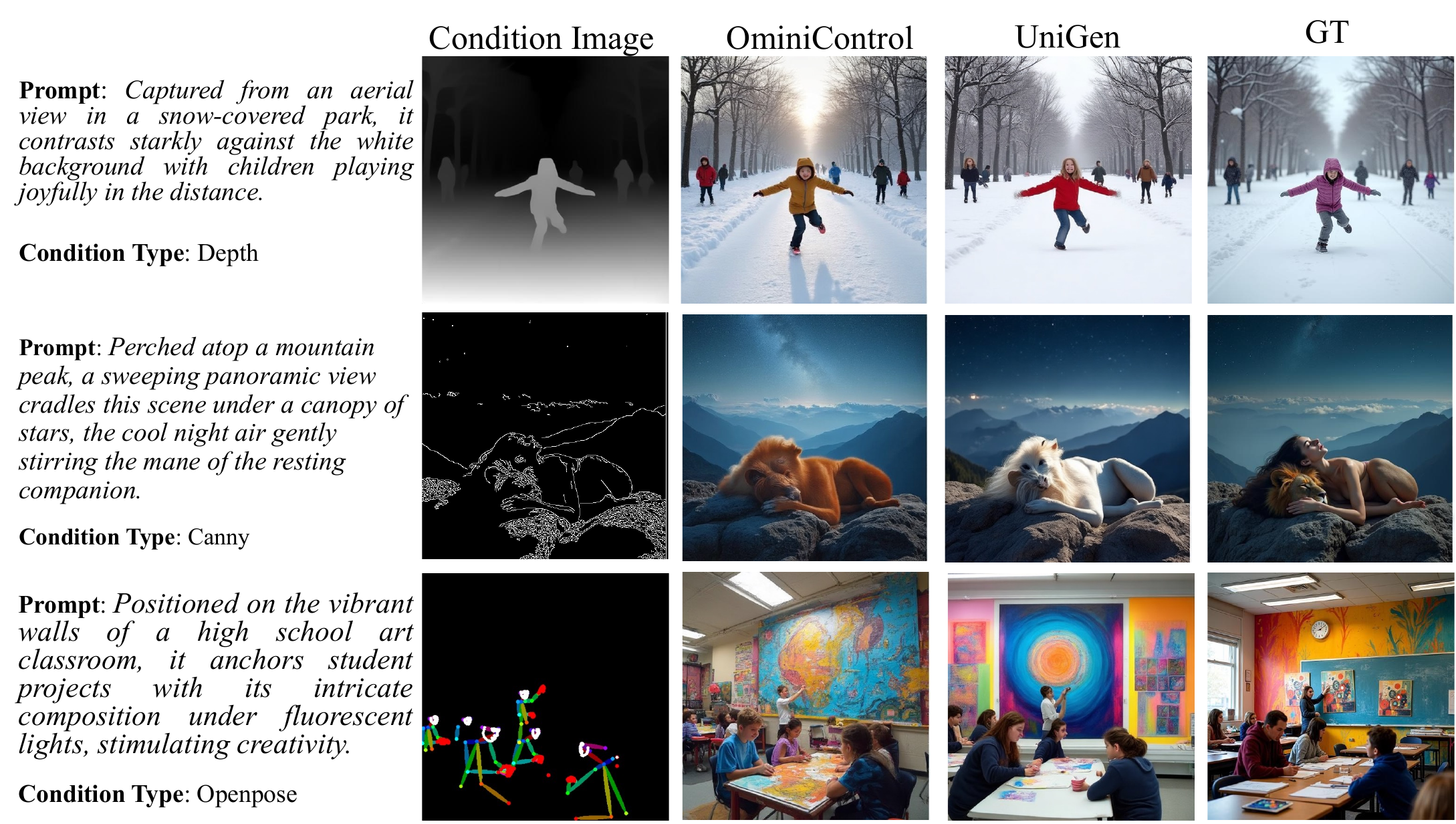}
    \caption{Visualization of failure cases for single-condition generation.}
    \label{fig:error_sample}
\end{figure}

Alongside visualizing high-quality samples, analyzing failure cases provides useful insights for improving the model. To this end, we select several unsuccessful examples under three types of conditions: Depth, Canny, and OpenPose. As illustrated in Fig. \ref{fig:error_sample}, two major issues can be observed: 1. Loss of local detail under strong spatial constraints.
As shown in the first row of Fig. \ref{fig:error_sample}, depth maps impose stronger spatial constraints than the other types of conditioning. Although this leads to better global structural consistency, it often results in blurred local details—for example, the child’s face or hands. This occurs because strong spatial constraints limit the model’s ability to generate fine-grained local information: the global structure is preserved, but the local details degrade. 2. Reduced spatial and semantic consistency under strong constraints. The third row of Fig. \ref{fig:error_sample} shows that Openpose enforces stronger local spatial constraints. However, this introduces a new issue: inconsistency between constrained regions and unconstrained surroundings. Although the human figures generated align well with the pose keypoints, their appearance becomes inconsistent with the environment. The figures appear detached from the background—more like “stickers”—resulting in a noticeably weakened general realism.

\subsubsection{Failure Case Analysis of Multi-Condition Controlable Image Generation}
\begin{figure}[h!]
    \centering
    \includegraphics[width=\linewidth]{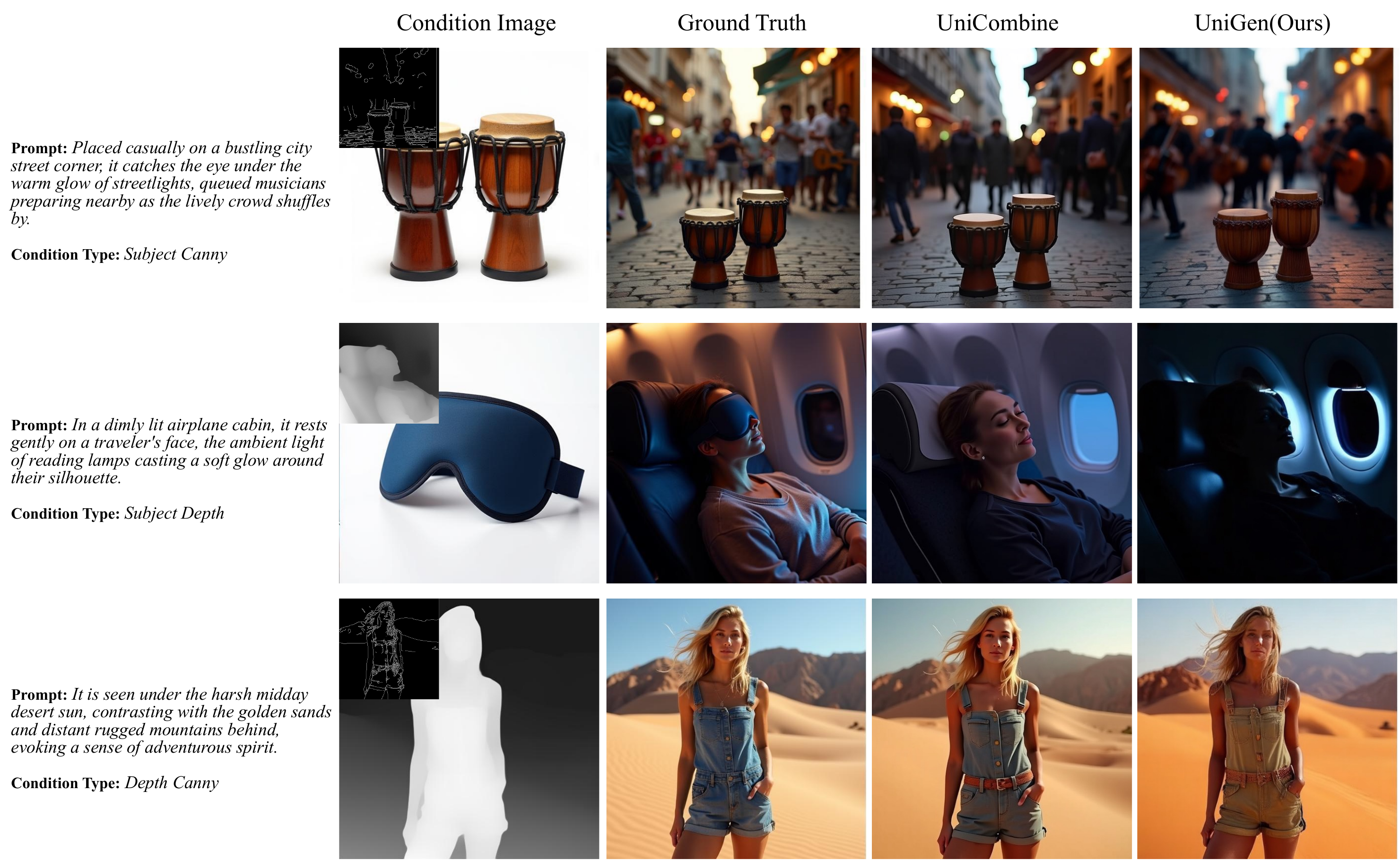}
    \caption{Visualization of failure cases for multi-condition generation.}
    \label{fig:error_sample_for_multicondition}
\end{figure}

To more comprehensively evaluate the model’s capabilities, we also present several failure cases under multi-condition control, which provide guidance and directions for future optimization. Specifically, in multi-condition tasks, the model is prone to condition confusion, inconsistent constraints, and even condition forgetting. For example, in the second row of Fig. \ref{fig:error_sample_for_multicondition}, although both our method and UniCombine effectively capture the spatial constraints specified by the Depth condition, a critical issue remains: the model forgets the Subject information, resulting in the subject not appearing in the generated image. We also observe that our generated images tend to have darker tones. This suggests that the model may incorrectly interpret the Subject information as a style cue, leading it to generate an overall dark, resting-state background. This insight indicates that stronger semantic guidance should be provided in multi-condition or subject-specific control tasks to prevent subject inconsistency.

Additionally, in the first row of Fig. \ref{fig:error_sample_for_multicondition}, although our method preserves the spatial layout consistent with the Canny condition, the subject undergoes partial modifications, introducing potential risks of unintended changes. Finally, under the combined Depth and Canny constraints, as shown in the third row of Fig. \ref{fig:error_sample_for_multicondition}, our method maintains spatial consistency better than UniCombine but still exhibits issues such as local blurriness in fine-grained details.

Through the analysis of failure cases, we identify three major limitations of our model: missing or blurred local details, inconsistencies in the main subject, and unintended alterations. These issues significantly constrain the model’s accuracy under multi-condition control. As our next step, we will address these problems by enhancing semantic conditioning to strengthen the constraint on subject information, thereby reducing condition confusion and preventing subject-related condition loss in multi-condition tasks.
In addition, we provide a detailed discussion of the limitations of the current approach and directions for future work in Section \ref{sec:limitation}.

\section{Conclusion}
\label{sec:conclusion}
This paper proposes a Unified Generative framework, UniGen, designed to handle various types of conditional image-guided generation tasks in a consistent manner. First, we introduce the Condition Modulated Expert (CoMoE) module, which enables unified processing of multiple condition types, effectively reducing computational and parameter redundancy caused by designing separate modules for each condition. Second, we present the WeaveNet architecture, which integrates global semantics from prompt guidance with local spatial information from conditional images, addressing semantic misalignment and enhancing both local consistency and overall image quality. Experimental results demonstrate that UniGen achieves superior performance across multiple datasets and evaluation metrics, validating its effectiveness and generality in image-to-image generation tasks.

\section{Limitations and Future Work}
\label{sec:limitation}
This paper proposes the Conditional Modulation Experts and WeaveNet architecture. With this approach, only a single copy of the backbone network is required to process conditional control information, while simultaneously supporting 12 different types of conditioning architectures. This significantly reduces computation and training costs for multi-condition tasks. However, our experiments also reveal several limitations.

First, the proposed method still relies on the duplication of independent branches of the backbone to process conditional information. Compared to LoRA-based approaches, the footprint of the resulting parameters remains relatively large. Second, analysis of the experimental results shows that the method weakens the influence of textual conditions to some extent, leading to inferior semantic alignment between generated outputs and textual prompts. We hypothesize that the step-by-step design of WeaveNet, while effective for handling conditional image inputs, may unintentionally diminish the strength of text-based constraints during training. To this end, our next steps will focus on improving the control of text-based conditioning, preventing the model from overemphasizing spatial alignment with the reference image at the cost of weakening semantic constraints. Specifically, we are investigating whether the semantic control strength can be dynamically adjusted during training based on the similarity between visual and semantic representations after condition injection. Our goal is to avoid semantic degradation caused by excessive reliance on spatial constraints.

In addition, most image-conditioned generation methods rely on the ControlNet-style condition injection mechanism, which performs element-wise feature addition. However, our results show that this approach may introduce a spatial-level semantic gap: the model tends to prioritize fitting the spatial structure provided by the condition image, leading to weakened global coherence. More concretely, this creates noticeable inconsistencies between conditioned and unconditioned spatial regions (see failure cases in Section \ref{sec:error_sample_analysis}), severely limiting both generation quality and practical usability. Therefore, we are exploring alternative condition injection strategies that maintain effective spatial constraints while improving overall image coherence. We have also conducted preliminary experiments with attention-based conditional injection. Although it enhances controllability, it significantly weakens spatial constraints. In future work, we will further investigate this issue and develop more robust solutions.

\section*{Acknowledgments}
This work was supported in part by the Fundamental Research Funds for the Central Universities 2025YJS051, in part by the National Natural Science Foundation of China under Grant 62473033 and 62576028, and in part by the Beijing Natural Science Foundation under Grant L231012.

\bibliographystyle{IEEEtran}
\bibliography{main}

\appendix
\begin{figure*}[h!]
    \centering
    \includegraphics[width=\linewidth]{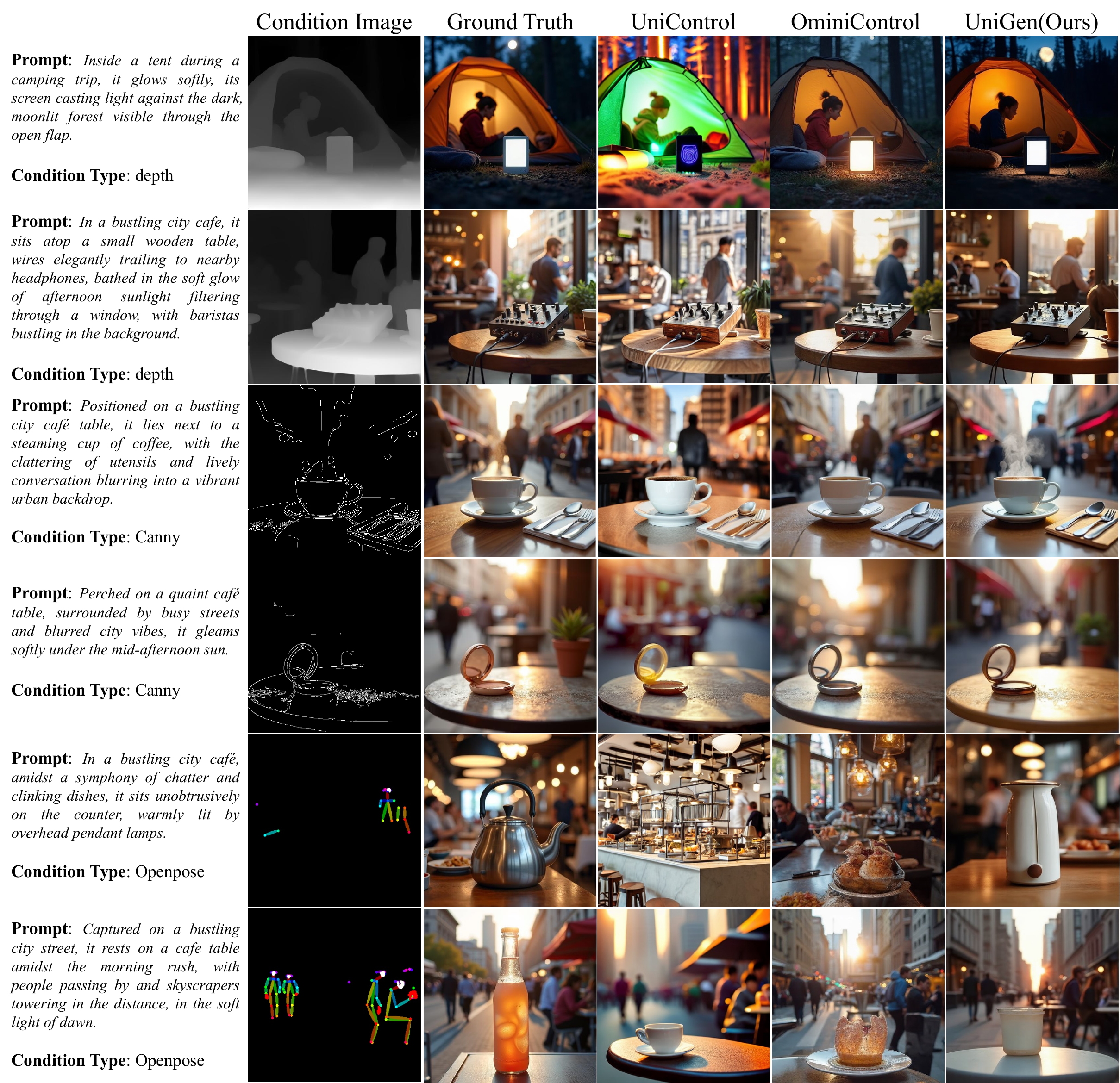}
    \caption{Based on the Subjects-200K dataset, we conduct a comprehensive comparison with UniControl \cite{Unicontrol} and  OminiControl \cite{ominicn} under Depth, Canny, and OpenPose conditions.}
    \label{fig:compare_omini_depth_canny_openpose}
\end{figure*} 
\begin{figure*}[h!]
    \centering
    \includegraphics[width=\linewidth]{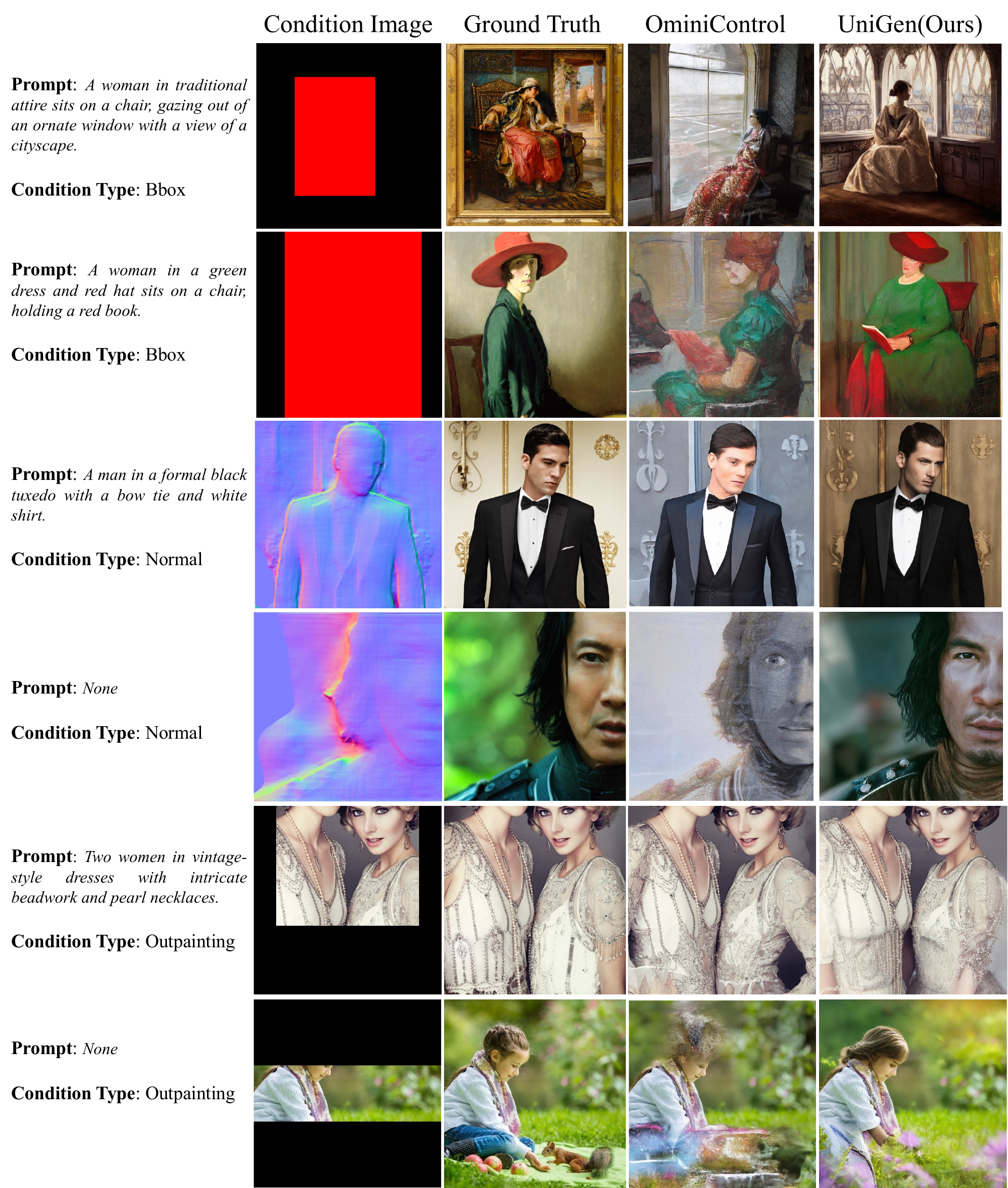}
    \caption{Based on the MultiGen-20M dataset, we conduct a comprehensive comparison with OminiControl \cite{ominicn} under Bbox, Normal, and Outpainting conditions.}
    \label{fig:compare_omini_bbox_normal_outpainting}
\end{figure*} 

\begin{figure*}[h!]
    \centering
    \includegraphics[width=\linewidth]{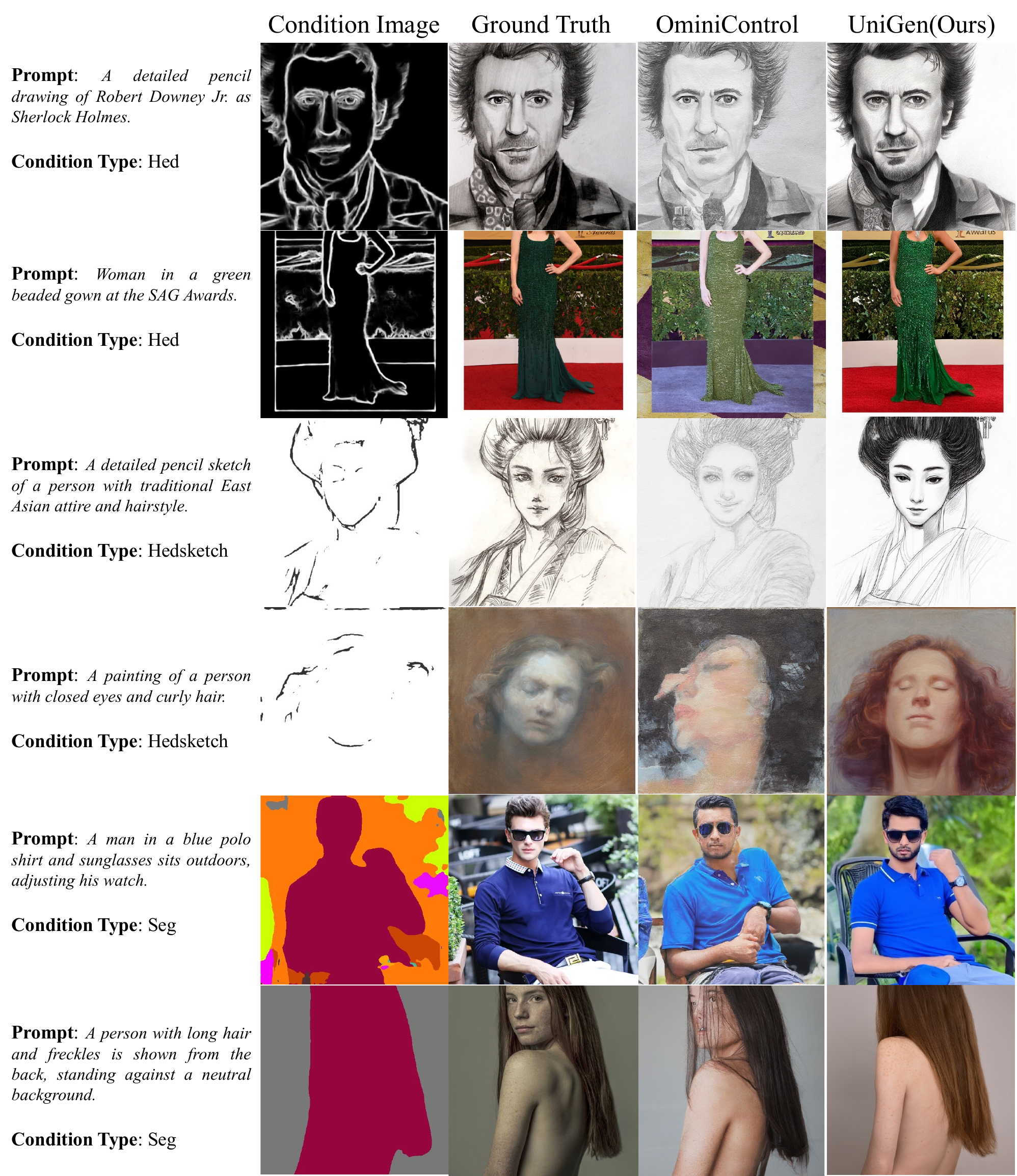}
    \caption{Based on the MultiGen-20M dataset, we conduct a comprehensive comparison with OminiControl \cite{ominicn} under Hed, Hedsketch, and Seg conditions.}
    \label{fig:compare_omini_hed_hedsketch_seg}
\end{figure*} 

\begin{figure*}[h!]
    \centering
    \includegraphics[width=\linewidth]{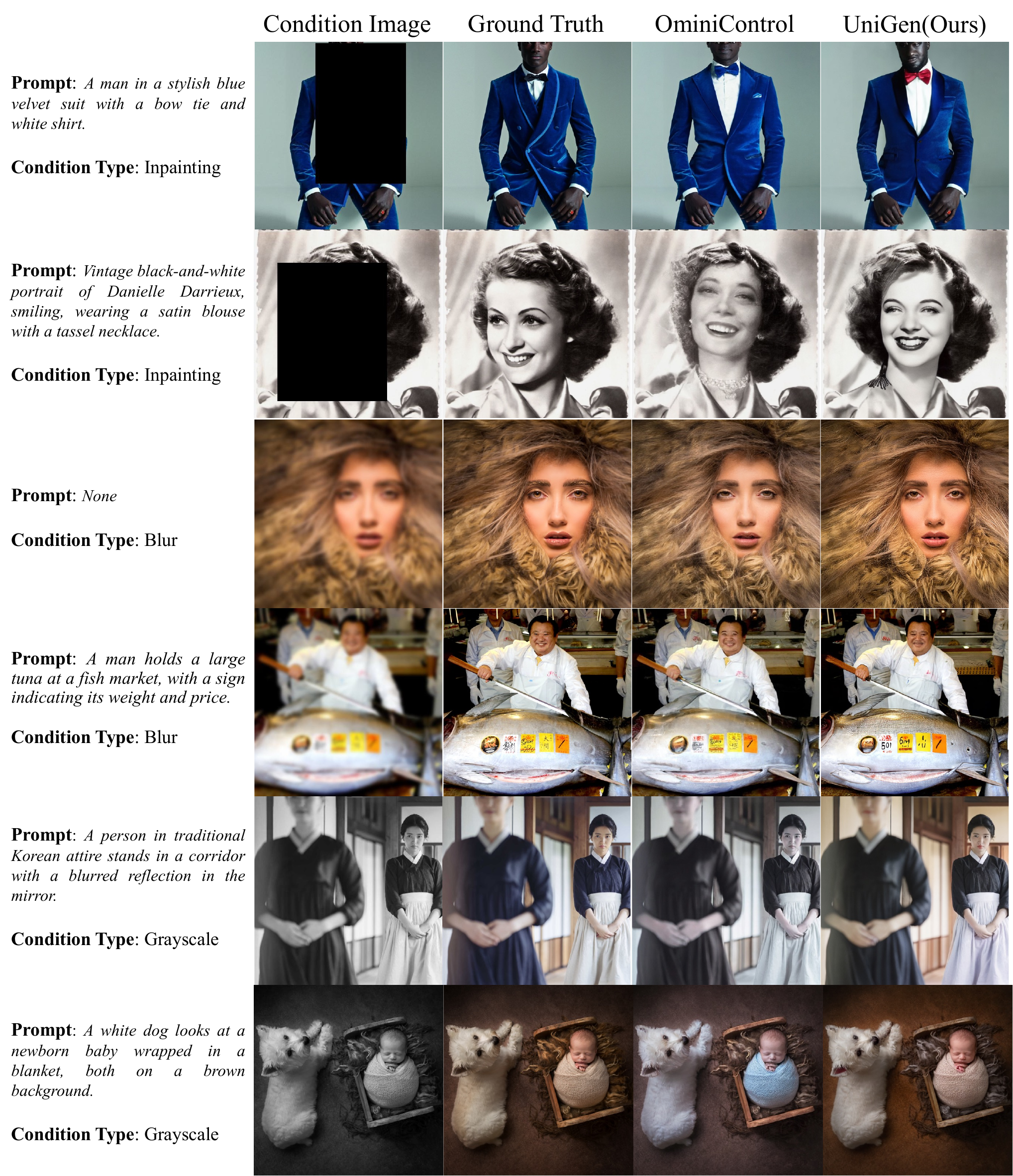}
    \caption{Based on the MultiGen-20M dataset, we conduct a comprehensive comparison with OminiControl \cite{ominicn} under Inpainting, Blur, and Grayscale conditions.}
    \label{fig:compare_omini_inpainting_blur_grayscale}
\end{figure*} 

\begin{figure*}[h!]
    \centering
    \includegraphics[width=\linewidth]{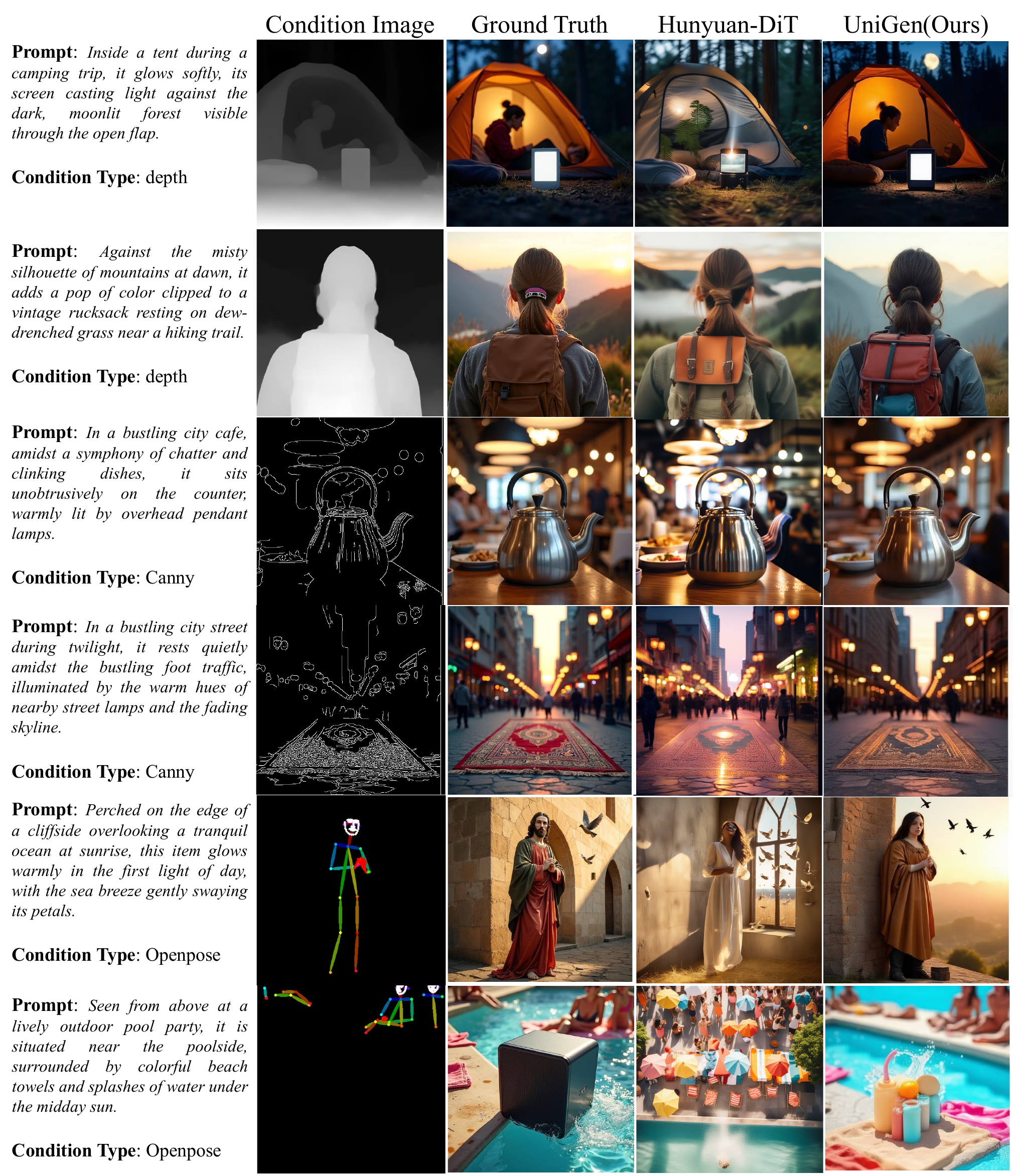}
    \caption{Based on the Subjects-200K dataset, we conduct a comprehensive comparison with HunyuanDiT \cite{hunyuan} under Depth, Canny, and OpenPose conditions.}
    \label{fig:compare_hunyuan_depth_canny_openpose}
\end{figure*}

\section{Comparative Analysis of Image Generation Quality}

Firstly, we conduct a visual comparative analysis of generation quality based on the Subjects-200K dataset. As shown in the Fig. \ref{fig:compare_omini_depth_canny_openpose}, we present results conditioned on Depth, Canny, and Openpose. In Fig. \ref{fig:compare_omini_depth_canny_openpose}, it is evident that under relatively simple conditions such as Depth and Canny, the difference in generation quality between OminiControl and our method is minimal. However, in finer details—for example, under Depth conditioning—OminiControl tends to overfit to the structural information of the condition image while neglecting the holistic semantic background described in the prompt.

In contrast, for the more challenging and sparse Openpose condition, OminiControl exhibits significant issues in local detail generation. For instance, in the last row, the spatial structure of the legs is incorrectly generated. We further extend the visual comparison to the MultiGen-20M dataset across nine additional conditioning types. As illustrated in Fig. \ref{fig:compare_omini_bbox_normal_outpainting}, under BBox, Normal, and Outpainting conditions, our method more effectively follows both the visual guidance from the condition input and the semantic cues from the prompt, producing more accurate and higher-quality results compared to OminiControl. Particularly in the Outpainting task, our method generates more precise extensions while maintaining global image consistency.

\begin{figure*}[h!]
    \centering
    \includegraphics[width=\linewidth]{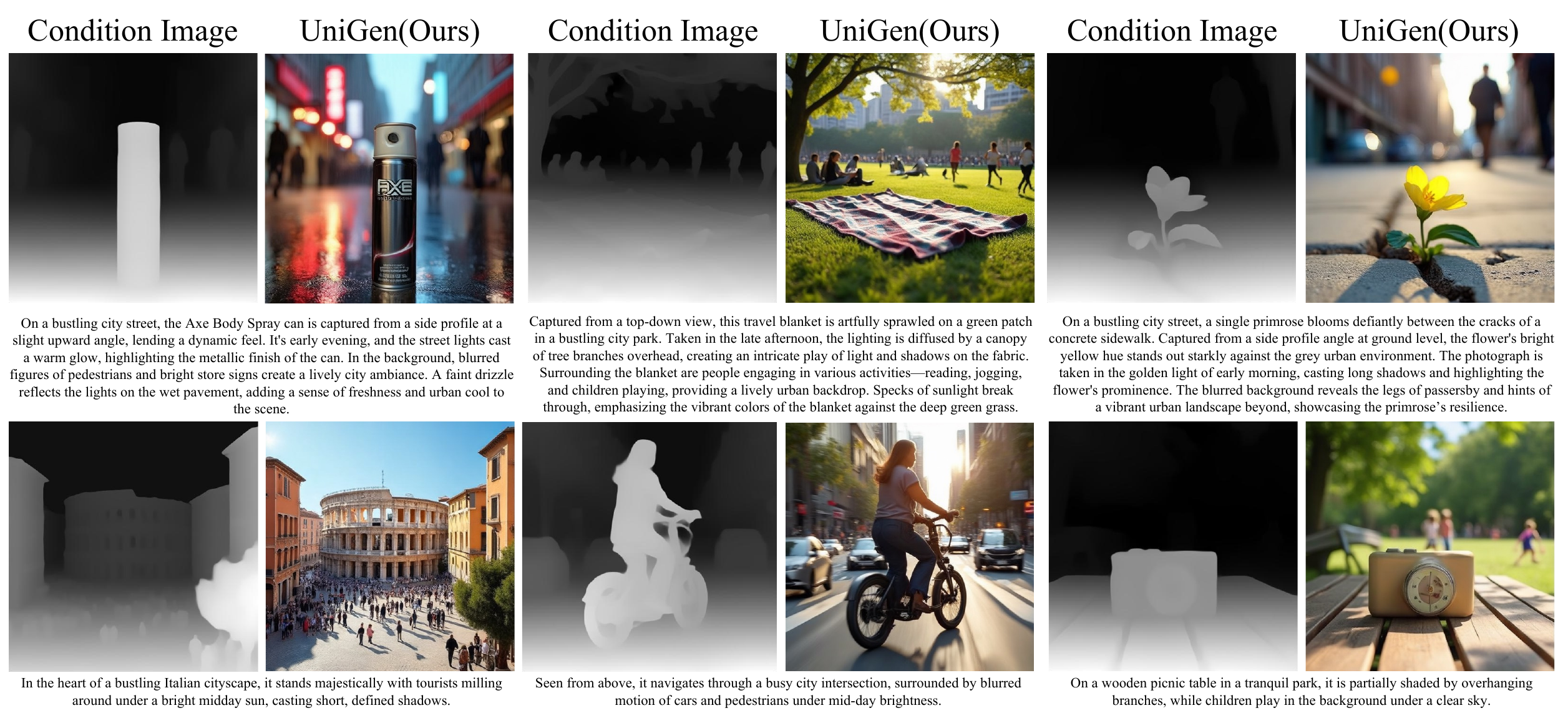}
    \caption{Generation results conditioned on Depth.}
    \label{fig:gen_depth}
\end{figure*} 

\begin{figure*}[h!]
    \centering
    \includegraphics[width=\linewidth]{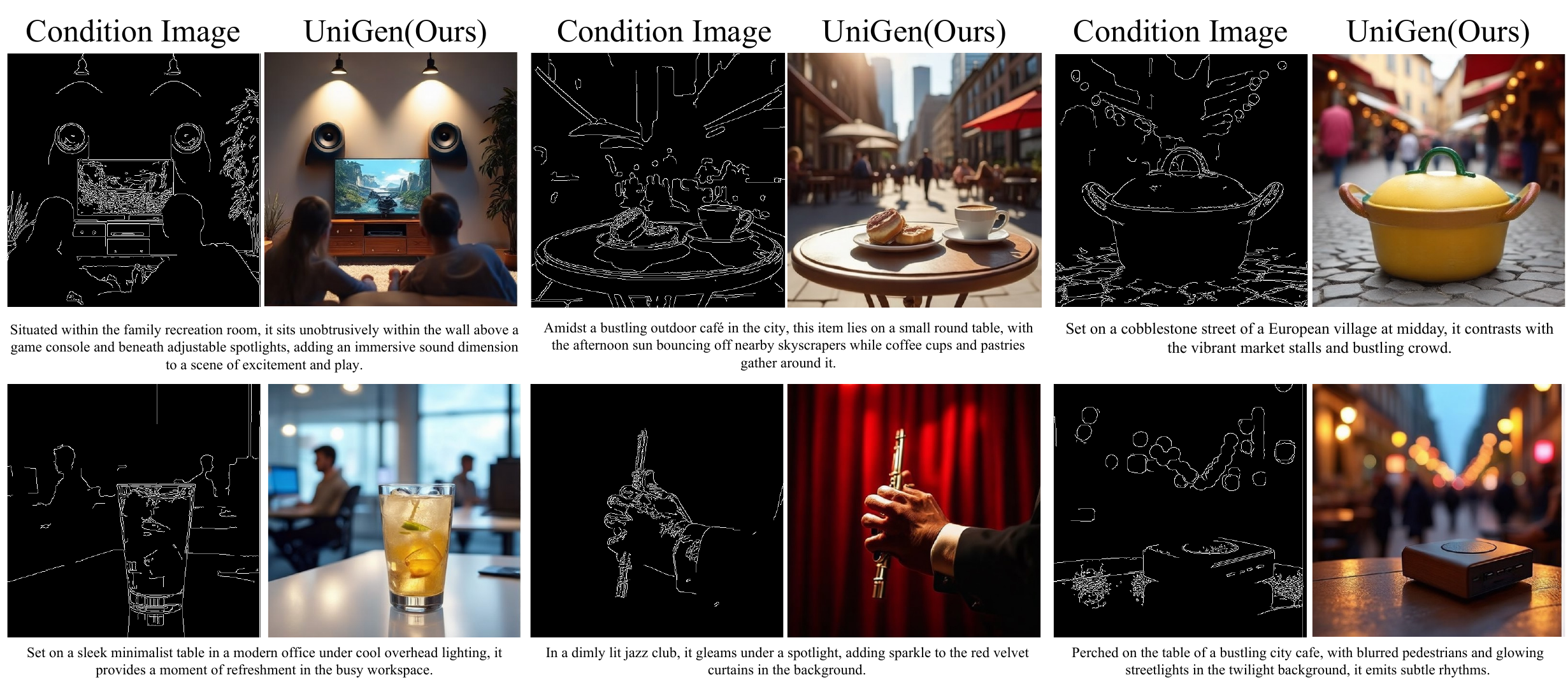}
    \caption{Generation results conditioned on Canny.}
    \label{fig:gen_canny}
\end{figure*} 

Furthermore, under the Hed, Hedsketch, and Seg conditions (Fig. \ref{fig:compare_omini_hed_hedsketch_seg}), the difference is less pronounced for relatively simpler conditions like Hed and Hedsketch. Nonetheless, our method provides richer detail. Notably, even under sparse guidance from Hedsketch, our model maintains strong instruction adherence and condition control, delivering more accurate and higher-quality results (e.g., fourth row). For the more complex and noisy Seg conditions, our method continues to produce accurate and high-quality images.

\begin{figure*}[h!]
    \centering
    \includegraphics[width=\linewidth,height=8.8cm]{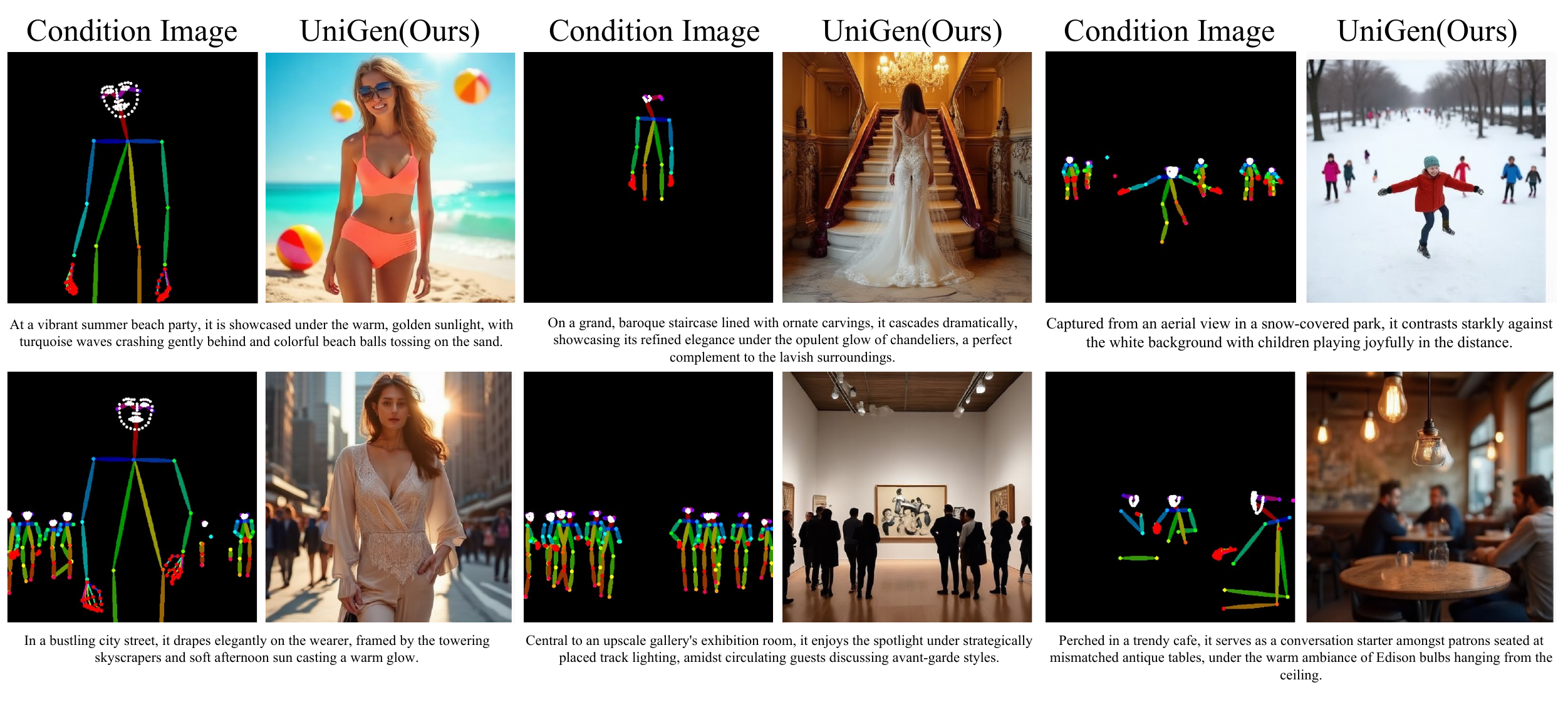}
    \caption{Generation results conditioned on Openpose.}
    \label{fig:gen_openpose}
\end{figure*} 

\begin{figure*}[h!]
    \centering
    \includegraphics[width=\linewidth]{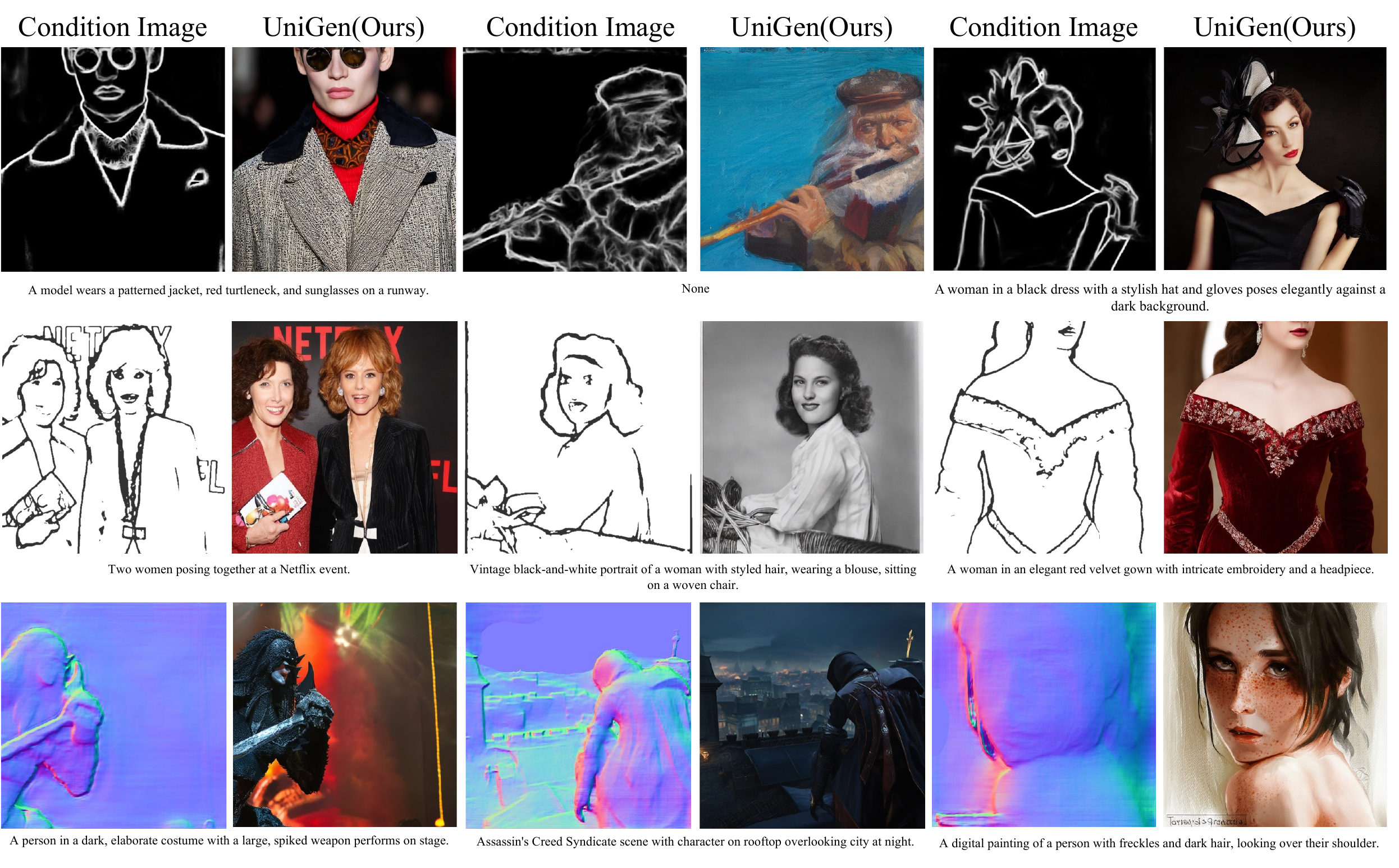}
    \caption{Generation results conditioned on HED, Hedsketch, and Normal.}
    \label{fig:gen_hed_hedsketch_normal}
\end{figure*} 

Finally, for Inpainting, Blur, and Grayscale tasks (Fig. \ref{fig:compare_omini_inpainting_blur_grayscale}), our method consistently generates more accurate and visually appealing results. Specifically, in the Inpainting task, the content generated within the masked region shows high consistency with surrounding context, yielding more natural and aesthetic outcomes. As Blur and Grayscale conditions typically offer more precise and dense control signals than other tasks, the generated outputs across different models appear generally similar in overall structure. However, differences emerge in fine-grained details and color fidelity. Under Blur conditions, our method generates sharper details, while under Grayscale conditions, it produces colorizations that are more realistic and aesthetically pleasing.

\begin{figure*}[h!]
    \centering
    \includegraphics[width=\linewidth]{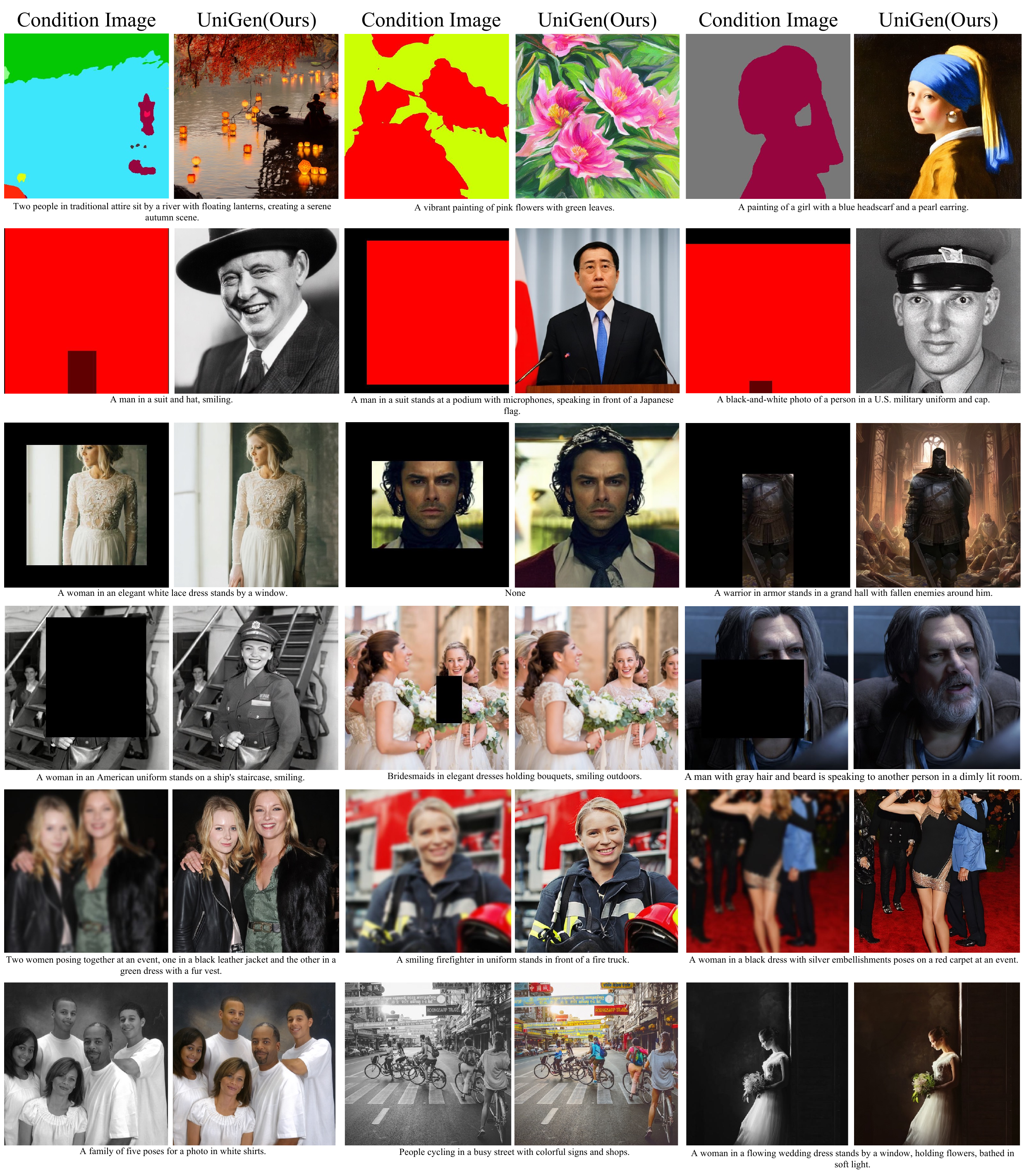}
    \caption{Generation results conditioned on Seg, Bbox, Outpainting, Inpainting, Blur and Grayscale.}
    \label{fig:gen_seg_bbox_outpainting}
\end{figure*}

To further assess the robustness of our method, we compare it with HunyuanDiT \cite{hunyuan}, a state-of-the-art open-source model. Using the official released weights \footnote{https://github.com/Tencent-Hunyuan/HunyuanDiT}, we evaluate conditional generation quality under Depth, Canny, and Openpose settings. As shown in Fig. \ref{fig:compare_hunyuan_depth_canny_openpose}, HunyuanDiT often suffers from loss of control signals. In sparse conditioning scenarios such as Openpose, it tends to either lose key information or overfit the condition input, resulting in reduced global consistency and the appearance of unrealistic "fake" elements, ultimately degrading the authenticity of the generated images.

\section{Visualization of Image Generation Results}
We present the generation results for all 12 conditions using the test sets of Subjects-200K and MultiGen-20M. As shown in Fig. \ref{fig:gen_depth} - \ref{fig:gen_seg_bbox_outpainting}, the generated images under various conditional images and textual instructions are illustrated. Overall, our method enables controllable image generation by effectively leveraging visual cues from conditional images and guidance from textual prompts, while ensuring high realism and visual quality.

\vfill

\end{document}